%% file: asplos23-opt-cc.tex
\documentclass[sigconf,screen,nonacm]{acmart}

\input{packages}
\input{commands}

\AtBeginDocument{%
  \providecommand\BibTeX{{%
    \normalfont B\kern-0.5em{\scshape i\kern-0.25em b}\kern-0.8em\TeX}}}

\setcopyright{acmcopyright}
\acmPrice{15.00}
\acmDOI{10.1145/3575693.3575712}
\acmYear{2023}
\copyrightyear{2023}
\acmSubmissionID{asplosb23main-p200-p}
\acmISBN{978-1-4503-9916-6/23/03}
\acmConference[ASPLOS '23]{Proceedings of the 28th ACM International Conference on Architectural Support for Programming Languages and Operating Systems, Volume 2}{March 25--29, 2023}{Vancouver, BC, Canada}
\acmBooktitle{Proceedings of the 28th ACM International Conference on Architectural Support for Programming Languages and Operating Systems, Volume 2 (ASPLOS '23), March 25--29, 2023, Vancouver, BC, Canada}
\received{2022-07-07}
\received[accepted]{2022-09-22}

\begin{document}

\title{\thiswork: Efficient Large NLP Model Training with 3D Parallelism Aware Communication Compression \vspace{5mm}}

\settopmatter{authorsperrow=4}
\author{Jaeyong Song}
\email{jaeyong.song@yonsei.ac.kr}
\orcid{0000-0001-9976-7487}
\affiliation{%
  \institution{Yonsei University}
  \city{Seoul}
  \country{South Korea}
}

\author{Jinkyu Yim}
\authornotemark[1]
\email{skyson00@snu.ac.kr}
\orcid{0000-0002-4719-6392}
\affiliation{%
  \institution{Seoul National University}
  \city{Seoul}
  \country{South Korea}
}

\author{Jaewon Jung}
\email{jungjaewon@yonsei.ac.kr}
\orcid{0000-0003-0770-9277}
\affiliation{%
  \institution{Yonsei University}
  \city{Seoul}
  \country{South Korea}
}

\author{Hongsun Jang}
\email{hongsun.jang@snu.ac.kr}
\orcid{0000-0003-4291-6124}
\affiliation{%
  \institution{Seoul National University}
  \city{Seoul}
  \country{South Korea}
}

\author{Hyung-Jin Kim}
\email{hj.windy.kim@samsung.com}
\orcid{0000-0002-2246-8830}
\affiliation{%
  \institution{Samsung Electronics}
  \city{Hwaseong}
  \country{South Korea}
}

\author{Youngsok Kim}
\email{youngsok@yonsei.ac.kr}
\orcid{0000-0002-1015-9969}
\affiliation{%
  \institution{Yonsei University}
  \city{Seoul}
  \country{South Korea}
}

\author{Jinho Lee}
\email{leejinho@snu.ac.kr}
\orcid{0000-0003-4010-6611}
\affiliation{%
  \institution{Seoul National University}
  \city{Seoul}
  \country{South Korea}
}


\thanks{This is an author preprint version of a paper which will appear in the proceedings of ASPLOS'23.\\
$^*$ Both authors contributed equally to this research. \\
$^{\dagger}$ Corresponding author.\\ \\ \\ }
\begin{abstract}
In training of modern large natural language processing (NLP) models, 
it has become a common practice to split models using 3D parallelism to multiple GPUs.
Such technique, however, suffers from a high overhead of inter-node communication.
Compressing the communication is one way to mitigate the overhead by reducing the inter-node traffic volume;
however, the existing compression techniques have critical limitations to be applied for NLP models with 3D parallelism in that 1) only the data parallelism traffic is targeted, and 2) the existing compression schemes already harm the model quality too much. 

In this paper, we present \thiswork, a fast and scalable distributed training framework for large NLP models with aggressive communication compression.
\thiswork differs from existing communication compression frameworks in the following ways:
First, we compress pipeline parallel (inter-stage) traffic. 
In specific, we compress the inter-stage backpropagation and the embedding synchronization in addition to the existing data-parallel traffic compression methods.
Second, we propose techniques to avoid the model quality drop that comes from the compression.
We further provide mathematical and empirical analyses to show that our techniques can successfully suppress the compression error.
Lastly, we analyze the pipeline and opt to selectively compress those traffic lying on the critical path. 
This further helps reduce the compression error.
We demonstrate our solution on a GPU cluster, and achieve superior speedup from the baseline state-of-the-art solutions for distributed training without sacrificing the model quality.
\end{abstract}

\ccsdesc[500]{Computing methodologies~Distributed computing \\ methodologies}
\ccsdesc[500]{Computing methodologies~Neural networks}

\keywords{Distributed Systems, Systems for Machine Learning, Large-scale NLP Training, Pipeline Parallelism, 3D Parallelism, Gradient Compression, Communication Optimization}



\maketitle

\vspace{3mm}
\section{Introduction}
\vspace{3mm}
In the era of deep learning (DL), the size of models has been growing at an exponential rate~\cite{gpipe, sc21efficient}.
Now, utilizing tens to hundreds of GPU-equipped nodes in a distributed manner to rapidly train a single model has become a common practice~\cite{megatron-lm, rajbhandari2020zero, trainingImagenetInHour}.
To achieve high speedups with multiple GPUs, the early work on distributed training employs \emph{data parallelism}~\cite{ddp, tictac, li2014scaling}.
With data parallelism, a DL model gets duplicated to multiple nodes with identical weight parameters.
Then, a batch of input data is split into the nodes and each node performs backpropagation on its copy of the DL model.
After every node completes backpropagation on their input data, inter-node communication is necessary for sharing the parameter gradients so that each node updates its copy of the DL model to have exactly identical states.
The volume of the communication is proportional to the size of the weights, which becomes a burden as the model size becomes larger.
Although the method has a problem with limited scalability of model size, it has been employed in many environments~\cite{flash,trainingImagenetInHour, minutes, minute15} because it is easy to implement.

To minimize the increase in the communication overhead, prior studies proposed to compress the parameter gradients of a DL model caused by data parallelism~\cite{dgc, terngrad, adacomp}.
The parameter gradients are known to be especially robust to some errors, so the DL model can tolerate a certain amount of misdirection.
By reducing the bitwidth of~\cite{terngrad}, taking the top-k of~\cite{dgc, scalecom, oktopk}, or performing low-rank approximation on~\cite{powersgd} the gradients, the parameter gradient compression techniques successfully reduce the inter-node traffic, making it feasible to perform distributed training on top of low-end ethernet~\cite{dgc}, and improve the speedup with high-speed inter-node networks~\cite{powersgd,scalecom}.
\newpage

However, with the rapid growth in the DL model sizes, especially for large natural language processing (NLP) models~\cite{bert, gpt2, brown2020language, turing-nlg}, a single GPU can no longer store a complete DL model even with the smallest batch size (i.e., one).
As a workaround, recent distributed training frameworks~\cite{megatron-lm, rajbhandari2020zero} suggest splitting the models into multiple pieces with 
\emph{pipeline} parallelism and \emph{tensor} parallelism.
Pipeline parallelism distributes the layers of a DL model to the nodes, whereas tensor parallelism partitions a layer into multiple sub-layers and distributes the sub-layers to the GPUs.
The three types (i.e., data, pipeline, tensor) of parallelism are collectively known as the \emph{3D parallelism} of distributed training.


Unfortunately, we found that existing compression methods targeting only the data-parallel traffic are inefficient for distributed training of a large DL model, especially for those using 3D parallelism~\cite{rajbhandari2020zero,megatron-lm}. 
First, data-parallel traffic is no longer the sole source of inter-node communication.
Pipeline parallelism requires point-to-point communication for passing forward activations and backward gradients between layers, and tensor parallelism incurs several all-reduce communications during forward and backward passes.
Second, we find that na\"{i}vely applying the existing compression techniques on recent larger NLP models causes a significant drop in the model quality (i.e., downstream task accuracy).
Moreover, applying the compression on the newly-introduced pipeline parallel traffic yields even more quality drops as illustrated in~\cref{sec:moti}. 
This states the need for new techniques that can suppress the compression error, or reduce the communication volume without loss. 
Last, existing compression methods~\cite{powersgd, scalecom, oktopk} overlook the opportunity coming from the pipelined schedule. 
When a training process is pipelined, much of the communication latency is hidden by the computation (i.e., forward and backward pass) of the following micro-batches. 
Thus, blindly compressing all communication traffic only yields more compression errors without any throughput gain.

\begin{figure}[t]
    \centering
    \includegraphics[width=0.97\columnwidth]{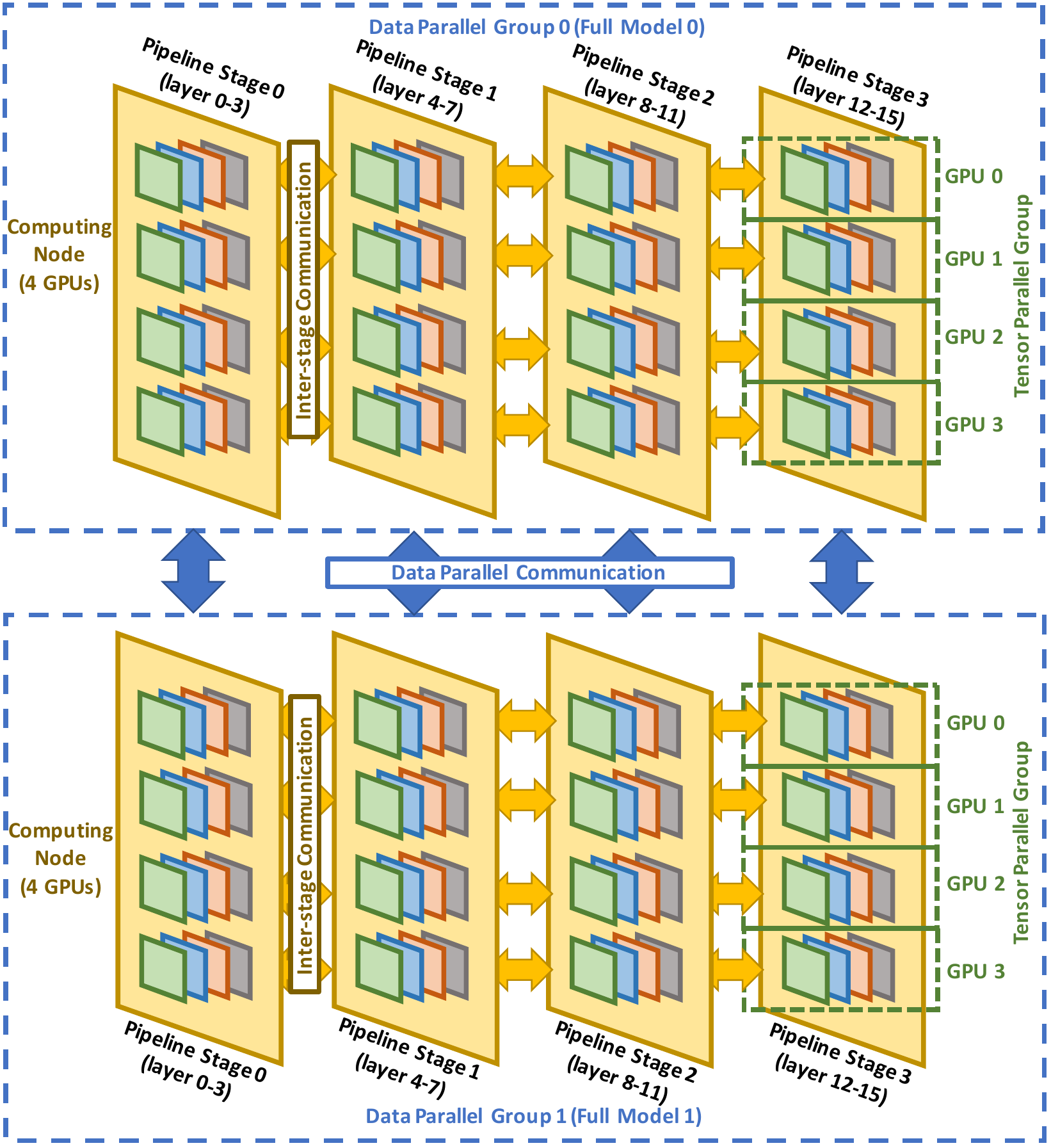}    \caption{Example distributed training configuration with 3D parallelism, with 2 data-parallel groups, 4 tensor-parallel groups, and 4 pipeline stages. \vspace{3mm}}
    \label{fig:background}
\end{figure}

In such circumstances, we propose \emph{\thiswork} (Compressed Communication), a fast and scalable distributed training framework for large NLP models.
Our goal is to achieve throughput gain by exploiting  characteristics of inter-node communication in 3D parallelism without sacrificing the model quality.
\thiswork employs the following three key ideas:
First, \emph{\ppmethod} targets the inter-stage communication of pipeline parallelism.
It compresses the inter-stage backward traffic which contains activation gradients.
By focusing on the pipeline epilogue and propagating compression errors in a lazy manner within an iteration, \ppmethod can increase speedup without compromising the model quality.
Second, \emph{\embmethod} fuses the two all-reduce communications from the embedding layers into a single all-reduce communication. 
The embedding synchronization occurs due to an embedding layer being shared at the beginning and the end of the network.
We find that fusing the synchronization reduces the communication volume without changing the mathematical outcome.
Third, \emph{\dpmethod} targets data parallel communication, but restricts the compression target to only a few stages.
The data parallel communication begins as soon as the backward pass of the corresponding stage finishes. Therefore, the earlier stages are likely to place the data parallel traffic on the critical path, so \dpmethod chooses not to compress some stages for less error.


We tested our approach on a cluster of 128 Nvidia A100 GPUs with 200Gb/s Infiniband HDR interconnect to achieve up to 15.09\% speedup on multi-billion NLP models without compromising application performance and up to 44.91\% speedup with comparable model quality.

Overall, our contributions can be summarized as follows:
\begin{itemize}
    \item We propose \thiswork, a fast and scalable distributed training framework which aggressively compresses inter-node communication while sustaining the model quality. To the best of our knowledge, \thiswork is the first work to accelerate large-scale NLP model training with inter-node communication compression.
    
    
    \item We propose three techniques tailored for reducing the communication volumes of 3D parallelism: \ppmethod, \embmethod, and \dpmethod. They significantly improve training throughput without suffering from the model quality drop.
    
    
    
    
    \item We demonstrate the high effectiveness of \thiswork by training two versions of GPT2~\cite{gpt2} models which have 8.3- and 2.5-billion-parameter respectively. In our large-scale GPU cluster setting equipped with a high-end interconnect, we obtain significant speedup on training.
\end{itemize}

\section{Background}

\subsection{3D Parallelism}
\label{sec:ppbackground}
3D parallelism is a strategy of utilizing data parallelism, tensor parallelism, and pipeline parallelism commonly adopted to large NLP model training~\cite{megatron-lm, rajbhandari2020zero} as depicted in~\cref{fig:background}.
\emph{Data parallelism}~\cite{DDL_sync_sgd, minute15, minutes} duplicates the same model weights on several groups. Then, mini-batches of a dataset are equally split into each data-parallel group.
After a forward and backward pass, each group $d$ has different parameter gradients $G^{(d)} = \nabla f^{(d)}(W)$ (where $f$ is a loss function and $W$ is the weight parameters) of the dedicated mini-batch for them. 
To maintain identical model weights in all the data-parallel groups, averaged parameter gradients ($\frac{1}{D}\sum^D_d G^{(d)}$) from all the $D$ groups are updated on weights. (i.e., $W \leftarrow W - \alpha\frac{1}{D}\sum^D_d G^{(d)}$).
It can reduce the training time because the forward and backward passes are parallelly executed in multiple GPUs.
The communication overhead of the averaged parameter gradient is the main cost issue for using data parallelism. 

\emph{Pipeline parallelism} and \emph{tensor parallelism}~\cite{krizhevsky2014one, chen2016training, chen2018efficient} let us handle large models that do not even fit in device memory by distributing a part of the model into multiple GPUs.
During forward and backward pass, these parallelisms should communicate activations and gradients between GPUs.
This mechanism accompanies huge communication volume, so the model should be carefully split to minimize the overhead.

Pipeline parallelism~\cite{gpipe, pipedream, pipemare} places a set of layers (i.e., a stage) to a GPU as depicted in~\cref{fig:background} and overlaps its executions in a pipelined manner as illustrated in Fig.~\ref{fig:timingbase}.
A number of prior work~\cite{gpipe, pipedream, pipemare} carefully schedule the executions such that the pipeline bubbles are minimized.
Between the stages, activations and activation gradients $\nabla f(Y)$ (where $Y$ is the intermediate activations) have to be communicated in a point-to-point manner for forward and backward passes.
\rev{Although the latency of many point-to-point communications are hidden by overlapping with computations, some communications are still not hidden, which become the target of this work. } 

Tensor parallelism splits a layer into multiple GPUs. 
By duplicating the activations to the GPUs in the same tensor-parallel group, each GPU applies different weight parameters to produce partial results. 
Hence, all-reduce communications are required during forward and backward passes for the activations and activation gradients.~\cite{megatron-lm} 


\begin{figure}
    \centering
    \includegraphics[width=0.97\columnwidth]{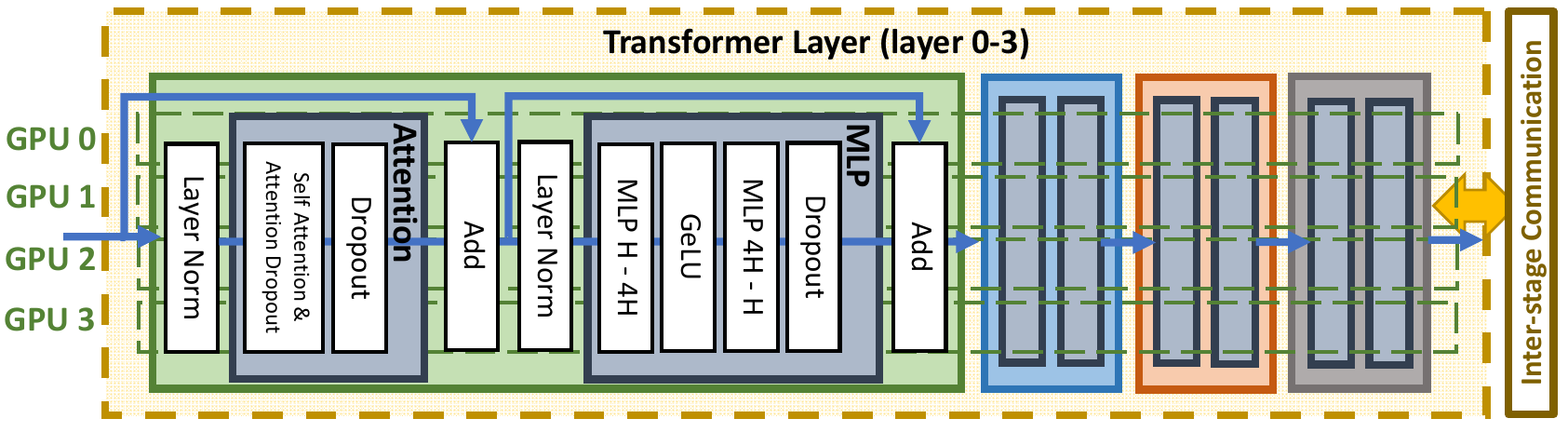}
    \caption{Layer structure of Megatron-LM. The weight parameters of each layer are split into multiple GPUs.}
    \label{fig:megatron}
\end{figure}

\subsection{Large-Scale NLP Model}
\label{sec:nlpmodelbackground}
In the NLP task, a large model~\cite{t5,bert,brown2020language} based on transformer~\cite{vaswani2017attention} is preferred due to its representational capability. 
Megatron-LM~\cite{megatron-lm, sc21efficient} is a framework for training extreme-scale NLP models using 3D parallelism.
\cref{fig:background} and~\cref{fig:megatron} show the details of Megatron-LM.
It applies \emph{tensor parallelism} to split model layers as depicted in Fig.~\ref{fig:megatron}.
To reduce the communication time, each tensor parallel group is placed within a server node such that its communications can utilize high-bandwidth intra-server interconnects (i.e., NVLink).

Multiple server nodes are utilized for data parallelism and pipeline parallelism. Megatron-LM also uses interleaved scheduling and 1F1B scheduling to reduce the pipeline bubbles and peak memory usage. 
Because the communications from data parallelism and pipeline parallelism take place in an inter-node network, those two communications often become bottlenecks in the 3D parallelism of Megatron-LM. 
This work targets reducing the volume of such communications, thereby maximizing the throughput of large NLP model training.

\subsection{Gradient Compression}
\label{sec:compression background}

Gradient compression is commonly used for mitigating the huge parameter gradient communication cost in data parallelism.
Top-k, quantization, and low-rank approximation are three popular approaches for gradient compression.

Top-k~\cite{dgc, MLSYS2021_8613985e} based approaches take top-k elements of the gradients per layer to compress.
In this case, top-k selection entails sorting overhead which is critical in fast training.
Some work~\cite{adacomp, scalecom} use quasi-sorting to reduce this overhead.
One problem with the top-k method is that as the number of GPUs in data parallelism increase, the total number of elements to communicate linearly increase, because each GPU will independently choose its own k elements. 
Furthermore, top-k methods require an additional gather operation for the chosen indices which induces more overhead than the actual parameter gradient sharing in a multi-worker environment.
ScaleCom~\cite{scalecom} resolves this gradient build-up problem because of the gather operation by using top-k index similarity between gradients in each worker.
Ok-Topk~\cite{oktopk} also successfully addresses these problems using its efficient top-k threshold estimation algorithm.

\rev{
Quantization-based approaches quantize gradients to reduce communications.
TernGrad~\cite{terngrad} uses ternary (-1, 0, 1) values to aggressively reduce communications.
AdaComp~\cite{adacomp} additionally combines residual addition and quantization to minimize the error from lossy compression.
SignSGD~\cite{signsgd} uses the sign of the momentum to significantly quantize gradients.
1-bit Adam~\cite{1bitadam} reduces the communication of the Adam optimizer by quantization using the stability of optimizer variables after warm-up iterations.
}

On the other hand, the low-rank approximation uses matrix factorization to reduce the total communication cost of gradients.
Factorization cost is the main bottleneck of matrix factorization.
PowerSGD~\cite{powersgd} diminishes this cost by iterating power-iteration, which is required for classical SVD, only once.
It reuses the factorized matrix from the previous gradient compression stage to minimize the error of compression.

One common drawback of these methods is that \rev{all the top-k, quantization, and low-rank} approximation methods are inherently lossy, and can degrade the accuracy of a model.
Many approaches try to mitigate this problem by considering the momentum~\cite{dgc}, providing feedback on errors~\cite{powersgd, adacomp}, or estimating the gradient movements~\cite{pipemare, pipescale}.
\thiswork is no exception, as it adopts a low-rank approximation to compress the communications. 
We propose several methods to maintain the model quality, while compressing the communications for the 3D parallelism.


\section{Motivational Study}
\label{sec:moti}
\begin{figure}[t]
    \centering
    \includegraphics[width=.85\columnwidth]{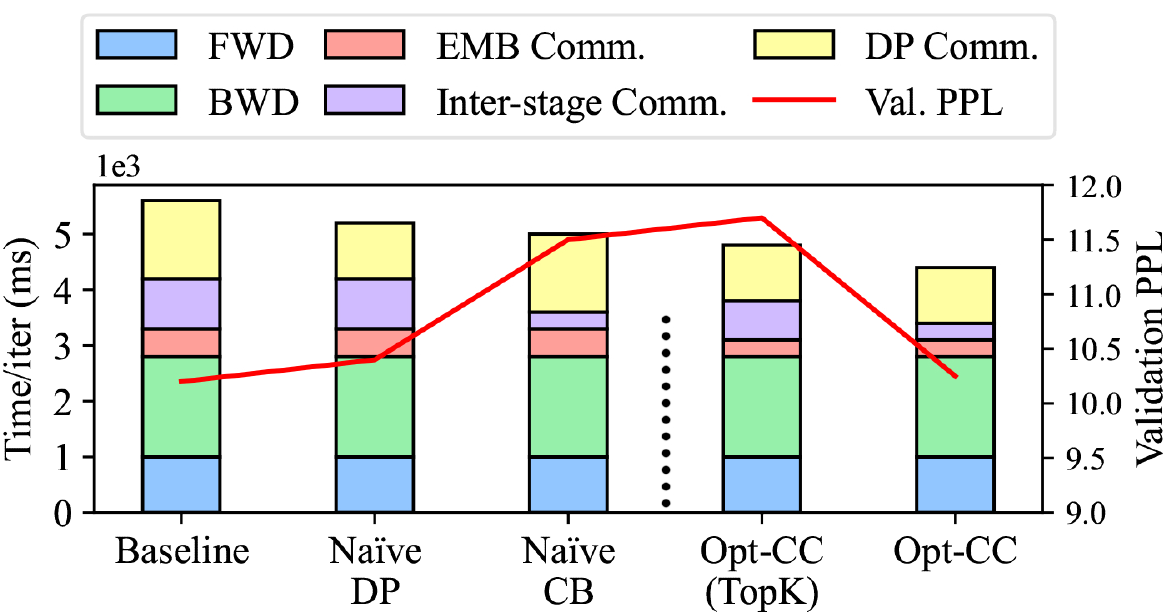}
    \caption{\rev{Motivational experiments. The breakdown demonstrates the overhead of communications, and naive compression yields severe increases in the validation perplexity.}}
    \label{fig:moti}
\end{figure}

\begin{figure*}[t]
    \centering
    \subfloat[Baseline.]{{ \includegraphics[width=0.99\textwidth]{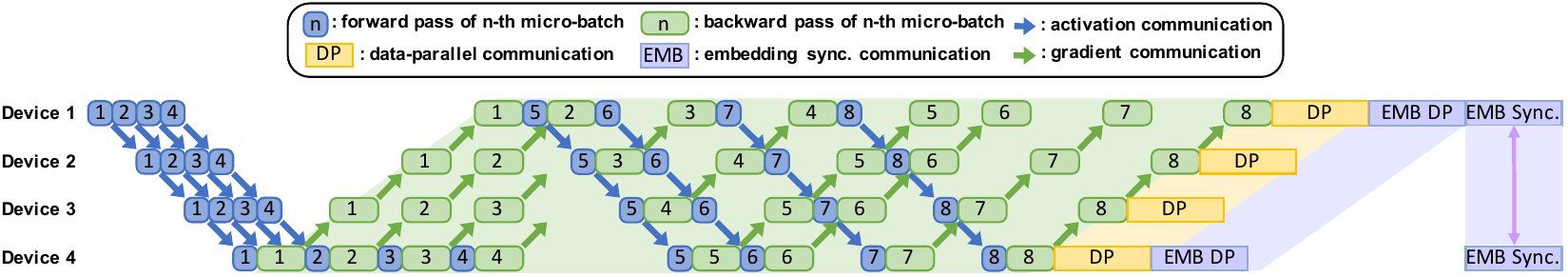}  }\label{fig:timingbase}}  \\
    \subfloat[Proposed \thiswork.]{{ \includegraphics[width=0.99\textwidth]{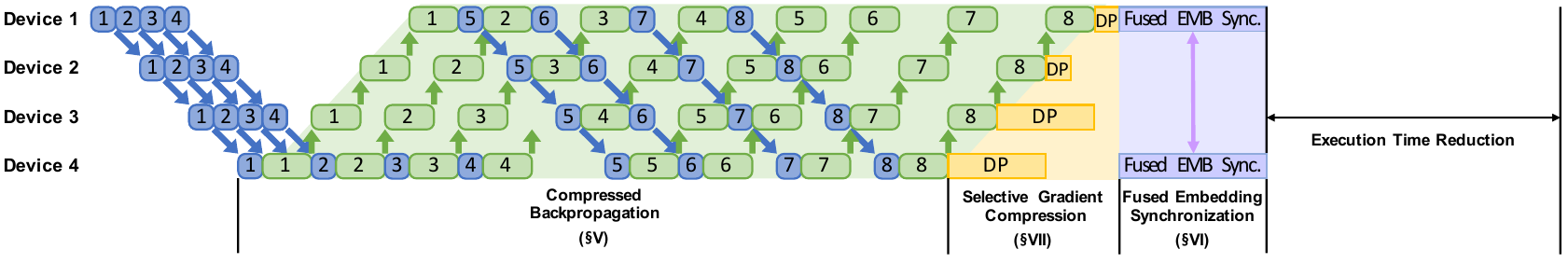}  }\label{fig:timingthis}} 
    \caption{Timing diagram of baseline and \thiswork, including the communication time between stages. Each stage and communication are not drawn as the exact scale for better visibility. 
    \label{fig:timing}}
\end{figure*}

\rev{\cref{fig:moti} illustrates the breakdown of 125K iterations training in a large NLP model (GPT-2.5B) using the popular Megatron-LM~\cite{megatron-lm} framework.
}
Because each GPU runs different stages with heterogeneous scheduling, we follow an approach similar to CPI stack~\cite{cpistack}. 
In other words, we turn off each communication/computation and observe the execution time difference.
The experiments have been done with 128 GPUs over 16 computing nodes connected with 200Gbps Infiniband HDR interconnect. 
Please refer to \cref{sec:environment} for a detailed setup.

Even with such high-speed interconnect, a significant portion is spent on the inter-node communication part (`DP Comm.' for data-parallel communication, `Inter-stage Comm.' for communication between pipeline stages, and `EMB Comm.' for embedding synchronization).
One exception is tensor-parallel all-reduce that happens intra-node, which we included in the FWD, and BWD bars.
In Megatron-LM, the tensor parallel GPUs are always placed in a single computing node, and connected via relatively fast NVLink with 600GBps bandwidth per GPU, which result in an almost negligible communication time.


\rev{Perplexity (PPL) is a representative validation metric in NLP, which measures how confidently the model predicts the next word after given sentences (the lower is the better).}
The `naive DP' bar depicts the training time and the validation PPL  when a low-rank gradient compression method~\cite{powersgd} is applied to the data-parallel communication to reduce its time.
\rev{It shortens the training time by reducing the volume of communication.}
However, contrary to the common wisdom on smaller models, even the modest amount of compression rate worsens the model quality as indicated by the increase in validation perplexity.
Considering that model quality is a key metric for DNN models, this much deterioration is not acceptable.
A similar observation can be made from the `naive CB' bar representing the compression of inter-stage communication.
We naively applied the same low-rank compression ratio to see the potential, even though no attempt has been made on the compression of inter-stage communication.
Similar to the observation in the data-parallel gradient compression, the compression yields an unacceptable rise in perplexity, and the phenomenon worsens when both communication types are compressed.

\thiswork targets applying compression to those inter-node communications, without sacrificing the quality of the model, as shown in the `Opt-CC' bar.
\thiswork compresses the communications until they only consume a negligible amount of execution time to achieve its speedup, and most importantly, maintains the perplexity and zero-shot task quality of the baseline method.
\rev{
As a result, \thiswork reduces the training time taken for 125K iterations from 8.00 days into 6.97 days, while maintaining the perplexity equal to that of the baseline.
Note that Opt-CC uses low-rank approximation for compression. 
We also depicted the result of top-k-based compression on the inter-stage communications in the `Opt-CC (TopK)' bar, but it brings worse perplexity because it is unsuitable for compressing point-to-point communications.
}

\section{Overview of \thiswork}

In this section, we describe three main components of \thiswork.
\cref{fig:timingbase} illustrates the baseline 1F1B scheduling~\cite{pipedream} of each pipeline stage, considering the communications.
The blue boxes represent the forward passes, and the blue arrows represent the inter-stage forward communication.
The green boxes represent backward passes where each pass takes approximately twice that of the forward passes, with the green arrows representing the inter-stage backward communications.
After the backward pass is complete, data-parallel communication takes place (\emph{DP}).
The communication is made with the GPUs of the same stage in other data-parallel ranks (yellow boxes in~\cref{fig:timing}).
\rev{Note that this also includes the communication for the parameters of embedding layers (\emph{EMB DP}), but it is depicted separately.}
Then, the embedding synchronization (\emph{EMB Sync}) happens, between the first stage and the last stage of the pipeline to sync weights because the first stage and the last stage share the same embedding weights.

The main goal of our proposed framework is to reduce the latest finish time or the whole execution, especially that of the first stage, because the next iteration starts from the forward pass of the first stage.
Our method pursues the goal by reducing the volume of communication in three sections shaded with green, yellow, and purple depicted in \cref{fig:timingthis}.

In the first technique, \emph{\ppmethod} (\cref{sec:ppmethod}), we compress the inter-stage communication. 
In the timing diagram, this technique has the effect of shaping the parallelogram-shaped green area closer to a rectangle, contributing to the shorter execution time.
We provide two enabler techniques, \emph{\lazy} and \emph{\epi}, which help avoid the model quality drop.
The second technique, \emph{\embmethod} (Section~\ref{sec:embmethod}) targets shared embedding layers, which generate two all-reduce communications. 
We fuse these two all-reduce communications into a single all-reduce communication for traffic volume reduction.
Lastly, \emph{\dpmethod} (Section~\ref{sec:dpmethod}) targets data-parallel traffic, but only compresses those lying on the critical path.
%
%
This provides a better trade-off between model quality and execution speed than traditional data-parallel traffic compression.

\section{\PPmethod}
\label{sec:ppmethod}

\begin{tcolorbox}
\begin{itemize}[leftmargin=*]
\item \textbf{Compression target}: Pipeline parallelism --- inter-stage backpropagation.
\item \textbf{Method}: Compress the inter-stage activation gradients, with the help of \lazy and \epi.
\end{itemize}
\end{tcolorbox}
With \ppmethod, we target the inter-stage backpropagation traffic (activation gradients, green arrows in \cref{fig:timingbase} and \ref{fig:timingthis}) using low-rank approximation.
Because the communication appears between computations of each stage, its performance impact becomes large, especially when there are many pipeline stages.
In principle, compressing the forward traffic could provide a similar speedup, but it would severely break the convergence of the model.
While naively compressing the backward traffic also breaks the convergence as demonstrated in \cref{fig:moti}, we provide two techniques for preserving the convergence, \lazy and \epi. 


\subsection{\LAZY}

\begin{figure}
\centering
\includegraphics[width=.85\columnwidth]{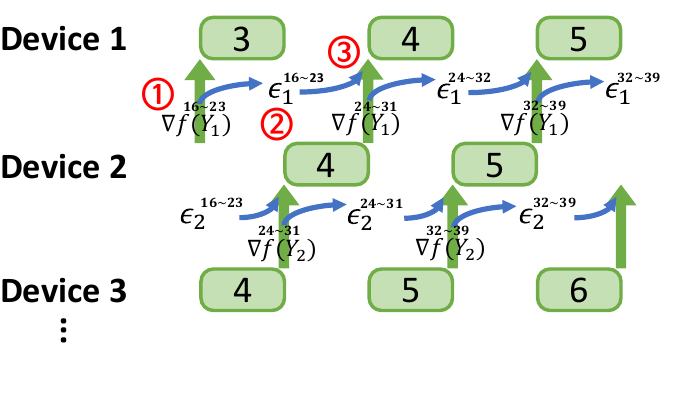}
\caption{\Lazy, with micro-batch size of 8.}
\label{fig:lazy_diag}
\end{figure}

\Lazy is a novel technique that enables \ppmethod without a model quality drop.
As shown in \cref{sec:moti}, naively applying compression to inter-stage backpropagation communication causes a severe model quality drop, due to errors from the lossy nature of the compression algorithm.

\cref{fig:lazy_diag} illustrates the inter-stage backpropagation with \lazy.
When \lazy is used, after compressed data are sent to the earlier stage for backpropagation, the error is preserved in the memory.
This preserved error is \rev{added to the backward traffic in the next micro-batch}.
For example, \circled{1} in the third micro-batch, the device 2 generates $\nabla f^{16-23}(Y_1)$: gradients from samples 16-23. \circled{2} the gradient is compressed and sent to device 1, while the compression error $\epsilon^{16-23}_1$ is preserved in memory. \circled{3} in the next micro-batch, when $\nabla f^{24-31}(Y_1)$ is generated, the $\epsilon^{16-23}_1$ is added. 
The sum is compressed before being sent, and the new error $\epsilon^{24-31}_1$ is preserved.

Even though the error is slightly delayed to the next micro-batch, 
the impact on the model quality is almost negligible because the model update only happens after all micro-batches are processed.
In other words, \lazy does not suffer from the weight staleness effect because all micro-batches are still executed based on the same version of the weights.
(See \cref{sec:evallazy} for the detailed model quality results.)

\rev{
\Lazy delays the error from a specific micro-batch to later micro-batches' errors.
For \lazy to work, the resulting average gradient should correctly approximate the average that would have resulted when compression is not used.
An intuition is that the gradients are being accumulated over the entire mini-batch, unlike the activations that directly affect the result of the model.
Because of this, we posit that the errors propagated in later micro-batch does not greatly affect the total sum. 
We found that this is true when the distribution of the errors are independent to that of the activations. 
In the following,
we provide a mathematical analysis of this condition. 
}

Consider a $K$-layer MLP\footnote{For brevity, we omit the bias terms and activation functions} as a toy example:
\begin{align}
    E = f (W_{K} \cdot W_{K-1}\cdots W_1 \cdot X^{(i)}), 
\end{align}
where $E$ is the error from some error function $f(\cdot)$, $W_{k}$ is the weights for layer $k$ and $X^{(i)}$ is the $i^{th}$ sample from a mini-batch.
In addition, we define $Y_k^{(i)}$ as the intermediate activation values at the output of layer $k$ from the $i^{th}$ sample of the mini-batch as below.
\begin{align}
    Y_k^{(i)} = W_k \cdot Y_{k-1}^{(i)}, \qquad Y_0^{(i)} = X^{(i)},  
\end{align}

Applying the common backpropagation, 
\begin{align}
    \nabla f^{(i)}(Y_k) &= W_{k+1} \cdot \nabla f^{(i)}(Y_{k+1}), \\ 
    \nabla f^{(i)}(W_k) &= \nabla f^{(i)}(Y_k) \cdot Y_{k-1}^{(i)},  
\end{align} 
Taking $k=K-1$ (the penultimate layer) for example, the parameter gradient for update becomes:
\begin{align}
    \nabla f^{(i)}(W_{K-1}) &= W_{K}\cdot\nabla f^{(i)}(Y_{K})\cdot Y_{K-2}^{(i)}, \label{eq:f}
\end{align}
Since the update occurs after the completion of the entire mini-batch,
assuming there are $N$ samples in a mini-batch, the update with SGD becomes:
\begin{align}
    W_k \leftarrow  W_k - \eta\cdot G_k, \\
    G_k = \frac{1}{N}\sum^N_i\nabla f^{(i)}(W_k),     \label{eq:g} 
\end{align}
where $\eta$ is a learning rate and $G$ represents the average gradient.
Now, consider each $W_k$ becomes an individual pipeline stage.
Then, the communicated backpropagation values between stage $k$ and $k+1$ are $\nabla f^{(i)}(Y_k)$.
When we apply compression, we essentially add an error vector $\epsilon_i^k$ to it (i.e., $\nabla f^{(i)}({Y_k})$ becomes $\nabla f^{(i)}({Y_k})+\epsilon_k^{(i)}$).
With \lazy, the errors are kept to be subtracted from the next micro-batch that has $n$ samples: 
\small
\begin{align}
    \label{eq:star2} \nabla f^{(i)}({Y_k}) &= W_{k+1} \cdot ((\nabla f^{(i)}({Y_{k+1}})-\epsilon_{k+1}^{(i-n)})+\epsilon_{k+1}^{(i)}), \\  
    \nabla f^{*(i)}({W_k}) &= ((\nabla f^{(i)}({Y_{k}})-\epsilon_k^{(i-n)})+\epsilon_k^{(i)}) \cdot Y_{k-1}^{(i)}, 
\end{align}
\normalsize

where $\nabla f^{*(i)}({W_k})$ is the approximate parameter gradient with \ppmethod.

Setting $k=K-1$ again,
\begin{equation}
\resizebox{.47\textwidth}{!}{%
$\begin{aligned}                
    \nabla f^{*(i)}(W_{K-1}) &= (\nabla f^{(i)}(Y_{K-1})-\epsilon_{K-1}^{(i-n)}+\epsilon_{K-1}^{(i)})\cdot Y_{K-2}^{(i)}, \\
                   &= (W_{K} \cdot (\nabla f^{(i)}(Y_K) - \epsilon_{K}^{(i-n)} + \epsilon_{K}^{(i)}) - \epsilon_{K-1}^{(i-n)} + \epsilon_{K-1}^{(i)})\cdot Y_{K-2}^{(i)}.  \notag
\end{aligned}$
}
\end{equation}

\normalsize
Aggregating them over $N$ samples to obtain $G_{K-1}^*$ yields, 
\begin{equation}
\resizebox{.48\textwidth}{!}{%
$\begin{aligned}
G_{K-1}^* &= \frac{1}{N}\sum^N_i\nabla f^{*(i)}(W_{K-1}), \label{eq:gstar}  \qquad\qquad\qquad\qquad\qquad\qquad  \\
 &= \frac{1}{N}\sum^N_i (W_{K} \cdot (\nabla f^{(i)}(Y_K) + \epsilon_{K}^{(i)}- \epsilon_{K}^{(i-n)})  + \epsilon_{K-1}^{(i)}- \epsilon_{K-1}^{(i-n)})\cdot Y_{K-2}^{(i)}
\end{aligned}$
}
\end{equation}

Substituting \cref{eq:f} results in the below equation.
\begin{equation}
\resizebox{.43\textwidth}{!}{%
$\begin{aligned}
          = G_{K-1} + \frac{1}{N}\sum^N_i (W_{K} \cdot (\epsilon_{K}^{(i)}-\epsilon_{K}^{(i-n)})  + \epsilon_{K-1}^{(i)}- \epsilon_{K-1}^{(i-n)})\cdot Y_{K-2}^{(i)}. \label{eq:errorsame} 
\end{aligned}$
}
\end{equation}

Restructuring the sums, 
\scriptsize
\begin{equation}
\begin{aligned}
          = G_{K-1} + \frac{1}{N}\sum^{N-n}_i (W_{K}\cdot \epsilon_K^{(i)} + \epsilon_{K-1}^{(i)})\cdot (Y_{K-2}^{(i)} - Y_{K-2}^{(i+n)})& \notag\\
          + \frac{1}{N}\sum^{n-1}_i 
          \Big\{ W_K \big(\epsilon_K^{(N-i)}\cdot Y_{K-2}^{(N-i)} - \epsilon_{K}^{(i-n)}\cdot Y_{K-2}^{(i)}\big)& \notag \\
    + \epsilon_{K-1}^{(N-i)} \cdot Y_{K-2}^{(N-i)}
          - \epsilon_{K-1}^{(i-n)} \cdot Y_{K-2}^{(i)}
          &\Big\}
                    \label{eq:restruct} 
\end{aligned}
\end{equation}

\normalsize

When $N \gg n$, the last term amortizes to yield
\begin{equation}
\resizebox{.40\textwidth}{!}{%
$\begin{aligned}
          \approx G_{K-1} + \frac{1}{N}\sum^{N-n}_i (W_{K}\cdot \epsilon_K^{(i)} + \epsilon_{K-1}^{(i)})\cdot (Y_{K-2}^{(i)} - Y_{K-2}^{(i+n)}).  \label{eq:actsame} 
\end{aligned}$
}
\end{equation}
For $G_{K-1}^*$ to correctly approximate $G_{K-1}$, the second term should be close to zero.
It is possible when one of the following two conditions suffices.
First,
    \begin{equation}
        \epsilon_k^{(i)} \approx \epsilon_k^{(i-n)}\qquad \forall k,i, \label{eq:errorcond} 
    \end{equation} 
    is derived from \cref{eq:errorsame}, which indicates the compression errors are always similar.
    Second,
    \begin{align}
            \epsilon_k^{(i)} \independent (Y_k^{(i)} - Y_k^{(i+n)}) \quad \forall k,i, \notag \\
            Avg(\epsilon_k^{(i)}) = 0, Avg(Y_k^{(i)} - Y_k^{(i+n)})=0 \quad \forall k,i, \label{eq:actcond}
    \end{align} 
    is derived from \cref{eq:actsame}. 

The first condition (\cref{eq:errorcond}) is intuitively difficult to be true, because it requires all errors to be almost equivalent to each other in their values.
Moreover, if it is true, compressing the forward activations with \lazy should also work.
However, in our experiments, compressing the forward inter-stage activation yielded divergence in the model, even with a very low compression rate.

On the other hand, the second condition (\cref{eq:actcond}) can be easily true, especially in the existence of batch/input normalization, which makes the average zero.
In \cref{sec:evallazy}, we will show that the second condition (\cref{eq:actcond}) is empirically true.
In practice, the error from the last micro-batch is lazily propagated in the first micro-batch of the last minibatch which further reduces the discrepancy.
Finally, please note that different $k$ values can be chosen to derive the same conditions, which we omit for brevity.

\subsection{Epilogue-Only Compression} 
\Epi is an additional technique in \ppmethod that provides better model quality without sacrificing the training speed. 
While compression techniques dramatically reduce the communication time, too much compression always has a risk of a drop in the model quality, which essentially becomes a trade-off between training speed and model quality.

\begin{figure}
    \centering
    \subfloat[Baseline epilogue.]{{ \includegraphics[width=0.75\columnwidth]{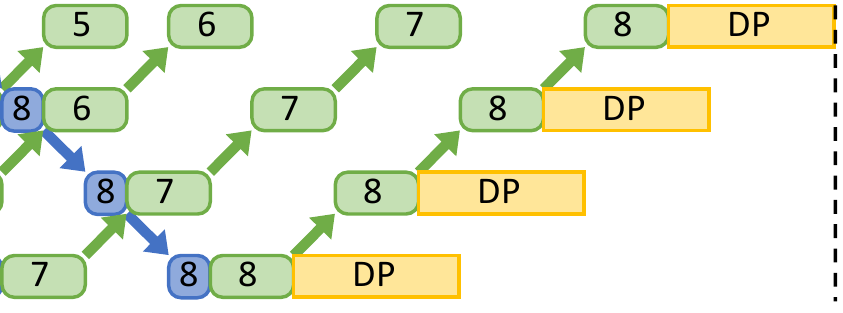}  }\label{fig:epilogue_before}}
    
    \subfloat[Compressed epilogue.]{{ \includegraphics[width=0.75\columnwidth]{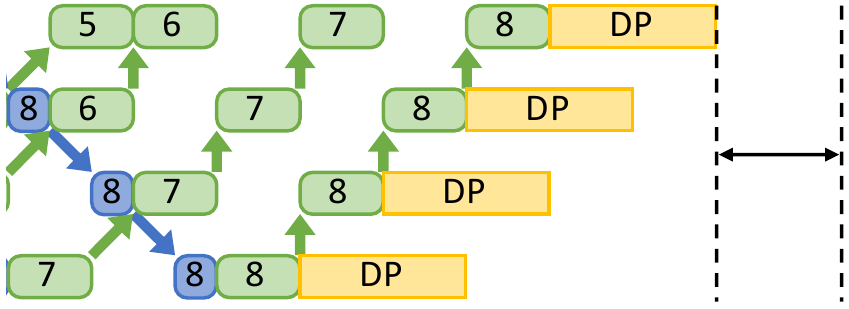}  }\label{fig:epilogue_after}} 
    \label{fig:epilogue}
    \caption{Illustration of \epi.}
\end{figure}

\rev{
Surprisingly, with pipeline parallelism, we can choose to compress the data that lie on the critical path.
Under the baseline scheduling shown in \cref{fig:timingbase}, many communications from inter-stage backpropagation are almost well-overlapped (hidden) with the computations. 
However, the communications from stages that lie on the critical path (called \emph{epilogue}) are not hidden by the computations, as depicted in~\cref{fig:epilogue_before}.}

With the above observation, \epi only compresses the communication in the epilogue part, as illustrated in~\cref{fig:epilogue_after}.
This causes less error in the training, and provides better quality to the model training.
As shown in \cref{fig:epilogue_after}, \epi does not reduce the speedup from the inter-stage communication compression when the inter-stage communication time is less than that of the corresponding stage's backward pass processing time. 
We found that this is true for high-end interconnects (i.e., $\geq$ 100Gbps), and find the \epi method useful.

\section{\EMBmethod}
\label{sec:embmethod}

\begin{tcolorbox}
\begin{itemize}[leftmargin=*]

\item \textbf{Compression target}: Pipeline parallelism --- embedding synchronization.
\item \textbf{Interacts with}: Data parallelism.
\item \textbf{Method}: Fuse two all-reduce communications of the shared embedding layer into a single all-reduce.

\end{itemize}
\end{tcolorbox}

When any of the weight parameters in the model are used multiple times in the network, the parameters aggregate gradients from multiple paths.
If the multiple paths lie within a single GPU, this does not incur any issue.
However, if they are executed on different GPUs (either with tensor parallelism or pipeline parallelism), 
they require synchronization of the parameter gradients, which is another type of communication. 

\begin{figure}
    \centering
    \subfloat[Baseline. Two all-reduce communications.]{{ \includegraphics[height=.44\columnwidth]{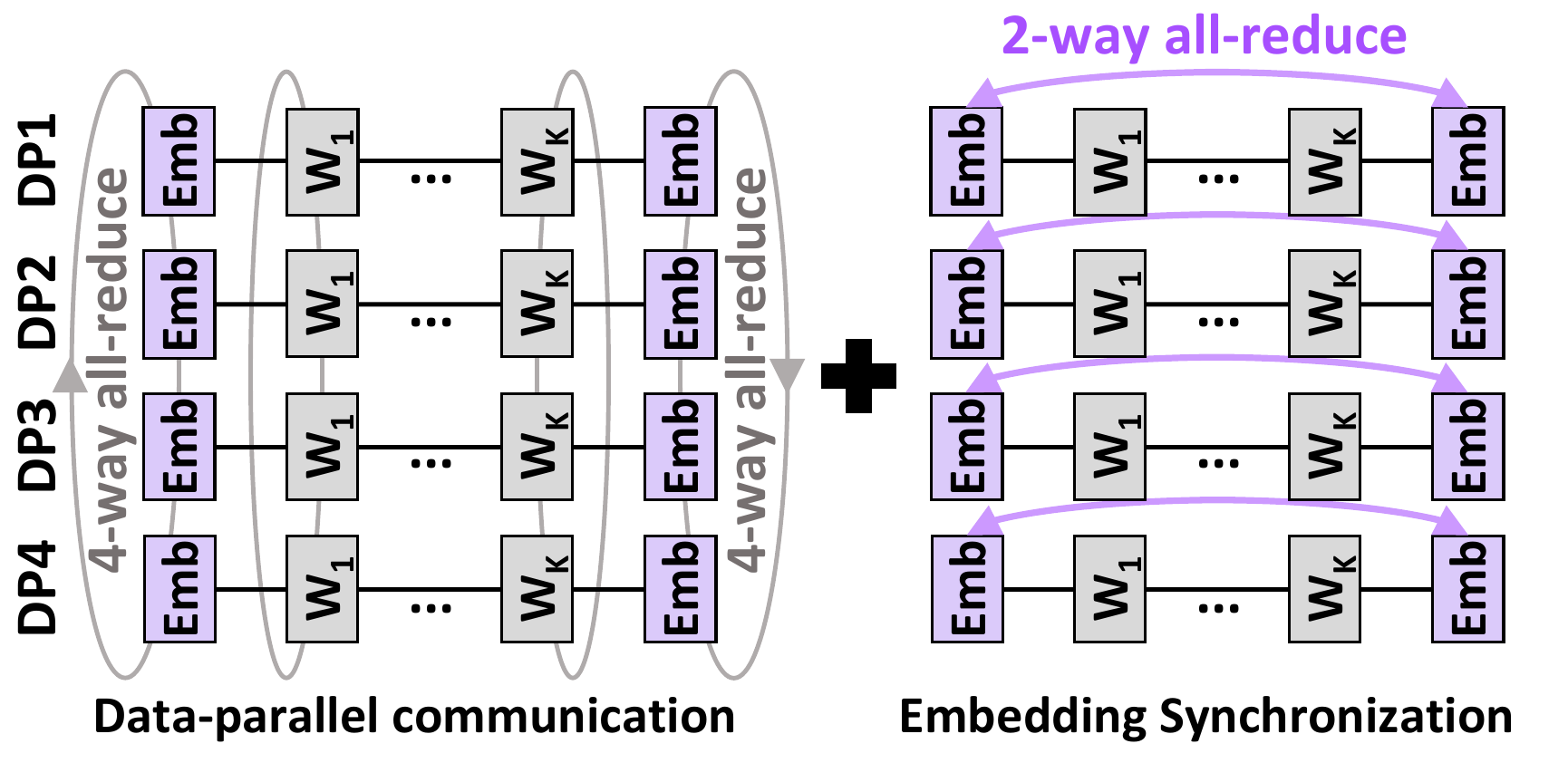}  }\label{fig:emb_before}}
    
    \subfloat[Fused embedding synchronization.]{{ \includegraphics[height=.41\columnwidth]{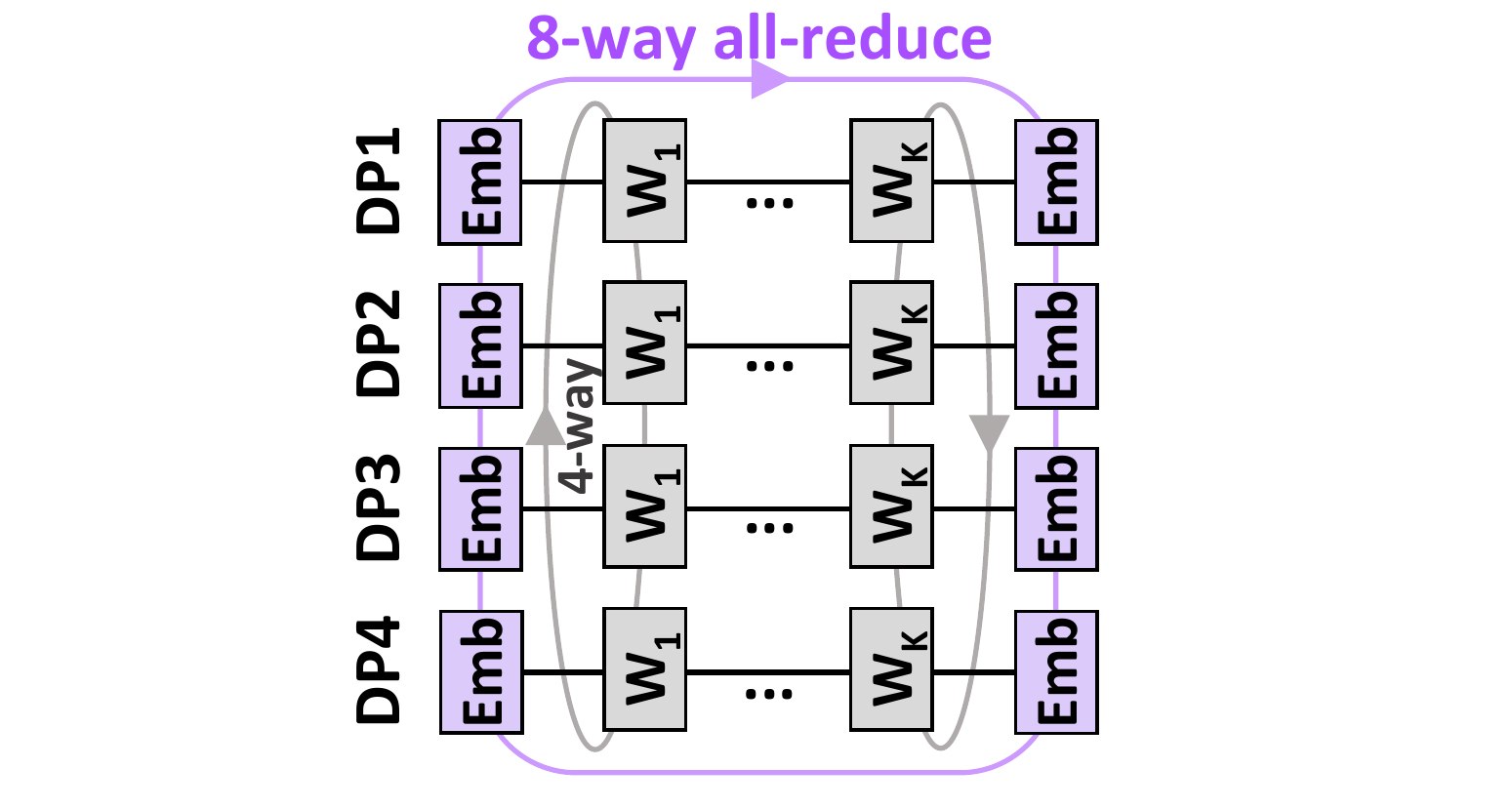}  }\label{fig:emb_after}} 
    \caption{Illustration of \embmethod.}    \label{fig:emb}
\end{figure}

In large NLP models that we primarily target, such a structure commonly appears with the embedding layers as depicted with purple boxes in~\cref{fig:emb}. 
If the output of the network takes a text format (as in the pretraining phase), the embedding layer is used twice: Once at the input for converting words into embedding values, and once more for converting the output vector into words.
Because they are at the beginning and at the end of the model, they always generate an inter-node communication if pipeline parallelism is used.
In existing solutions shown in \cref{fig:emb_before}, the embedding layer is duplicated in both the first and last stages. 
After the layer-wise gradients are collected from data parallel communication, the duplicated embedding layers share the gradients in an all-reduce pattern in a separate communication phase.


Effectively, the functionality of the embedding synchronization is to share the gradients, identical to that of the data-parallel communication (i.e., all-reduce). 
Thus, we can fuse the two all-reduce communications associated with the embedding layers (one from the data-parallel ways and the other from embedding synchronization) into a single all-reduce as illustrated in~\cref{fig:emb_after}.
It is known that, for $R$ ranks participating in an all-reduce communication for communication volume $V$, the cost is $2V\cdot\nicefrac{(R-1)}{R}$~\cite{optimizationOfCollectiveComm}.
Because the number of ranks for embedding synchronization is always two, the conventional cost for the embedding layer $C_{Emb}$ becomes
\begin{align}
C_{Emb}=2V\cdot\frac{D-1}{D} + 2V\cdot\frac{1}{2} = V\cdot\frac{3D-2}{D},
\end{align}
where $D$ is the number of data-parallel groups.
With \embmethod, the number of ranks for the fused synchronization becomes $2\cdot D$, and the cost becomes
\begin{align}
C_{Emb\_fused}=V\cdot\frac{2D-1}{D}.
\end{align}
\rev{Thus, as $D$ becomes large, the improvement approaches 50\% over the baseline embedding synchronization time.}
For $D=4$ used in our settings, the theoretical benefit already reaches 42.9\%.


\section{\DPmethod} 
\label{sec:dpmethod}

\begin{tcolorbox}
\begin{itemize}[leftmargin=*]

\item \textbf{Compression target}: Data parallelism.
\item \textbf{Interacts with}: Pipeline parallelism.
\item \textbf{Method}: Compress the current bottleneck stages for the trade-off between speed and the model quality.

\end{itemize}
\end{tcolorbox}

\begin{figure}
    \centering
    \includegraphics[width=\columnwidth]{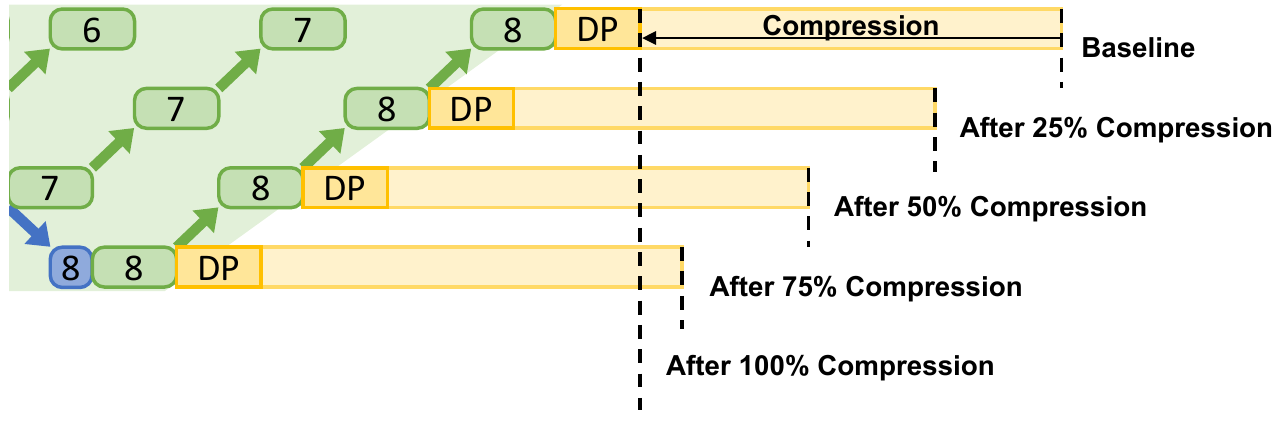}
    \caption{\rev{Illustration of \dpmethod.}}
    \label{fig:dpmethod}
\end{figure}

In \thiswork, we provide \dpmethod as an optional technique to obtain a trade-off between model quality and speed when compressing data-parallel traffic.
As shown in \cref{fig:moti}, compression of data-parallel traffic results in a big drop in the model quality, making the technique unacceptable despite its large benefit in the execution time.
Even with a low compression rate, the drop is still severe.
We believe the reason comes from the weight staleness issue~\cite{gpipe}.
While the error from compression can be sent to the next iteration~\cite{powersgd, scalecom}, unlike \ppmethod, the error is applied after the weight update, and has a stale effect on the weight.

Instead of relying on the compression rate, \dpmethod provides a better knob by considering the pipeline schedules into account for the stage selection.
In conventional data parallel-only distributed training, the gradient exchange communication for data parallelism happens after the backward propagation, and lies on the critical path.
However, because of pipeline parallelism, the later stages finish early from the backpropagation and they can start the communication before others.
Then, selectively compressing the stages, starting from the earlier stages, can adjust how much data we compress and therefore how much quality we lose.
For example, in \cref{fig:dpmethod}, the rightmost edge of the light-yellow box on stage 1 represents the finishing time for uncompressed data-parallel communication.
When the first stage is compressed, the new bottleneck becomes stage 2, and further compressing the traffic from stage 2 yields the new bottleneck to stage 3.
This scheme provides a much better trade-off, as we will demonstrate in \cref{sec:evalselcomp}.


\section{Implementation}

\thiswork has been implemented over the publicly available Megatron-LM code~\cite{megatrongithub}.
The interleaved pipeline scheduling~\cite{megatron-lm} has been applied to reduce the scheduling bubbles.
For the low-rank approximation, we adopted PowerSGD~\cite{powersgd} implementation.
\rev{We used the low-rank approximation for both \ppmethod and data-parallel gradient compression, because top-k methods are not suitable for point-to-point communications as shown in~\cref{sec:moti}, and~\cite{powersgd} shows good performance compared to other compression methods~\cite{mlsysgrad}.}
All of our additional implementations have been made using native PyTorch~\cite{pytorch} 1.8 APIs, such that it does not require external libraries or recompilation of PyTorch.

For the \ppmethod, we wrote a custom low-rank compression code to support point-to-point communication, 
and integrated it into the methods of Megatron-LM's \texttt{p2p\_communications.py}.
To realize \lazy, private variables were declared in the \texttt{PowerSVD} class to store the errors between micro-batches.
For \epi, \texttt{schedule.py} was modified to apply compression on the epilogue part of the communications.
For the \dpmethod, we inherited the \texttt{DistributedDataParallel} class 
and overrode the \texttt{allreduce\_gradients()} method.
%
For the \embmethod, we again modified the \texttt{allreduce\_gradients()} function.
We detected the embedding layer by searching for the word \emph{word\_embeddings} in the name of the layer, 
and replaced the communication with the custom method.




\begin{table}[]
\footnotesize
\centering
\caption{Experimental Environment}\label{tab:environment}
\begin{tabular}{cccc}
\toprule
\multirow{6}{*}{\textbf {HW}} & \multirow{4}{*}{\makecell{Server\\Node}} & \#Nodes & 16 \\
&& CPU & 2$\times$EPYC 7543, 32 cores \\
&& Memory & 1TB DDR4 ECC \\
&& GPU & 8$\times$ Nvidia A100 \\
\cmidrule(lr){3-4}
&\multirow{2}{*}{Interconnect} & Intra-node & NVLink (600GBps / GPU)\\
&& Inter-node & Infiniband HDR (200Gbps) \\

\midrule

\multirow{9}{*}{\textbf {Model}} & \multirow{3}{*}{Common} & Micro-batch  & 8 \\
&& Total mini-batch & 512 \\
&& \#iterations & 230K\\

\cmidrule(lr){3-4}
&\multirow{3}{*}{\makecell{GPT-8.3B}} & \#layers &  72\\
&& Hidden dim. & 3072 \\
&& Ways & TP8 / DP4 / PP4 \\
\cmidrule(lr){3-4}
&\multirow{3}{*}{\makecell{GPT-2.5B}} & \#layers &  52\\
&& Hidden dim. & 1920 \\
&& Ways & TP8 / DP4 / PP4 \\

 \bottomrule
\end{tabular}
\end{table}

\begin{table*}[t]
\centering
\caption{Pretraining (230K iterations) training time speedup and validation set perplexity using 128 GPUs.}
\begin{tabular}{clcccc}
\toprule
                    &    & \textbf{Baseline} & \textbf{CB (Speedup)} & \textbf{CB+FE (Speedup)} &  \textbf{CB+FE+SC (Speedup)} \\ \midrule
\multirow{2}{*}{\textbf{GPT-8.3B}}
& Training Time & 37.27 days & 34.83 days ($+$7.01\%) & 32.84 days ($+$13.49\%) & \textbf{25.72 days ($+$44.91\%)} \\ 
&Val. Perplexity& \textbf{8.10} & \textbf{8.10} & \textbf{8.10} & 8.20 \\\midrule
\multirow{2}{*}{\textbf{GPT-2.5B}}
& Training Time & 14.72 days & 13.63 days ($+$8.00\%) & 12.79 days ($+$15.09\%) & \textbf{12.55 days ($+$17.29\%)} \\ &Val. Perplexity& \textbf{9.31} & \textbf{9.31} & \textbf{9.31} & 9.55 \\\bottomrule
\end{tabular}
\label{tab:pretrain}
\end{table*}

\section{Evaluation}
\label{sec:eval}

\subsection{Experimental Environment and Method} 
\label{sec:environment}

We conducted our experiments based on the environments listed in \cref{tab:environment}.
All experiments have been performed under a fixed set of nodes, such that unexpected variations could be minimized.
Following \cite{megatron-lm},
we pretrained GPT with 8.3B parameters (GPT-8.3B) and GPT with 2.5B parameters (GPT-2.5B) model to figure out the convergence of the model when using the proposed methods. 
The GPT-8.3B has 72 layers and a hidden dimension size of 3072, while the GPT-2.5B model has 52 layers, with a hidden dimension size of 1920.
To utilize 128 GPUs, both models had eight tensor parallel groups that match the number of GPUs within a node, four data-parallel groups, and four pipeline stages unless otherwise stated.
We pretrained models for 230K iterations, where the baseline model reaches the LAMBADA~\cite{paperno2016lambada} task accuracy reported in~\cite{megatron-lm}. All perplexity data are the result of training for 230K iterations unless otherwise stated.

Following \cite{megatron-lm}, we executed pretraining of chosen NLP models with RealNews~\cite{realnews}, Wikipedia~\cite{bert},  CC-stories~\cite{cc_stories} and OpenWebtext~\cite{openwebtext} datasets. 
We concatenated all these datasets and created a corpus. 
The datasets were preprocessed using the original Megatron-LM code~\cite{megatrongithub}, including the elimination of short documents and deduplication.
\rev{For validation metrics, we also followed~\cite{megatron-lm} (using 5\% of the dataset as a validation set) including the holdout, splitting documents into training and validation at the beginning.}

For the number of ranks in low-rank approximation-based compression, we used 128 for data-parallel gradient compression and 16 for \ppmethod unless otherwise stated, following the settings of the transformer-based model (around 10$\times$ compression) in~\cite{powersgd}.
\rev{We followed~\cite{powersgd} because compression algorithms are well-studied research areas.
We empirically chose 75\% stage compression for \dpmethod.} 
We ran 30K of warm-up iterations for all models, also following the practice from~\cite{powersgd}.











\subsection{Training Performance} 
\label{sec:evalpretrain}

\begin{figure}
    \centering
    \includegraphics[width=0.8\columnwidth]{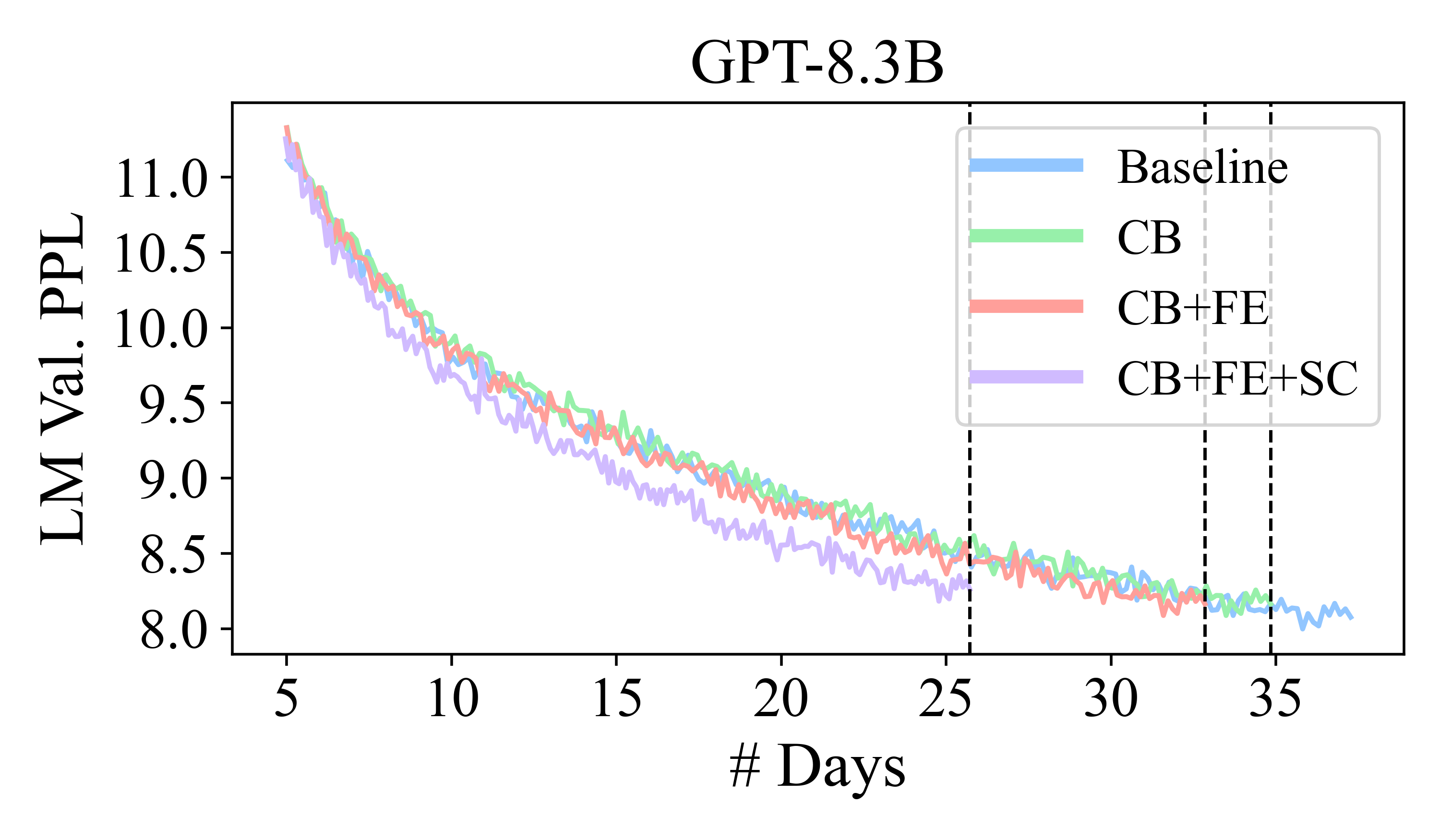}
    \caption{Pretraining validation LM perplexity of the proposed methods and the baseline.}
    \label{fig:pretrain}
\end{figure}




\begin{table*}[t]
\centering
\caption{\rev{Accuracies on zero-shot tasks which indicates the expressibility of pretrained models.}}
 \begin{tabular}{lccccccccc}
\toprule

                    \multirow{2}{*}{\textbf{Tasks}}  & \multicolumn{4}{c}{\textbf{GPT-8.3B}} & \multicolumn{4}{c}{\textbf{GPT-2.5B}} \\
                    \cmidrule(lr){2-5}\cmidrule(lr){6-9}
                    & \textbf{Baseline} & \textbf{CB} & \textbf{CB+FE} &  \textbf{CB+FE+SC}  & \textbf{Baseline} & \textbf{CB} & \textbf{CB+FE} &  \textbf{CB+FE+SC} \\ \midrule
 LAMBADA~\cite{paperno2016lambada} & \textbf{66.82\%} & 66.35\% & 66.35\% & \underline{65.79\%}  & \textbf{62.00\%} & 61.93\% &  61.93\% & \underline{61.15\%}  \\ 
 PIQA~\cite{PIQA} & \textbf{75.52\%} & \underline{74.05\%} & \underline{74.05\%} & 74.27\% & \underline{71.76\%}  & \textbf{72.63\%} &  \textbf{72.63\%} & 71.93\%\\ 
 MathQA~\cite{2019mathqa} & \textbf{24.36\%} & 23.55\% & 23.55\% & \underline{23.52\%}  & 24.15\%& \textbf{24.25\%} & \textbf{24.25\%} & \underline{23.42\%}  \\
 WinoGrande~\cite{winogrande} & \underline{63.22\%} & \textbf{63.30\%} & \textbf{63.30\%} & \underline{63.22\%}   & \textbf{62.19\%} & \underline{60.62\%} & \underline{60.62\%} & 61.33\% \\
 RACE~\cite{lai2017race} & \textbf{37.89\%} & \textbf{37.89\%} & \textbf{37.89\%} & \underline{37.32\%}   & \underline{33.88\%} & \textbf{35.12\%} & \textbf{35.12\%} & 34.64\%\\ \bottomrule

 \end{tabular}
\vspace{-2mm}
\label{tab:zeroshot}
\end{table*}

\cref{tab:pretrain} show the training speed, and the validation perplexity of \thiswork, on the chosen models. 
In the table, `Baseline' refers to the original Megatron-LM without any communication compression.
In addition, `CB' refers to \ppmethod where the \lazy and \epi are used together, `FE' refers to \embmethod, and `SC' refers to \dpmethod.
On an 8.3B parameter model, \thiswork achieves 44.91\% speedup over the baseline (no compression) with marginal perplexity increase on CB+FE+SC, or
13.49\% speedup without compromising perplexity on CB+FE.
A similar trend can be seen from the 2.5B model, with a 17.29\% speedup with a small perplexity increase on CB+FE+SC or a 15.09\% speedup without a perplexity increase on CB+FE.

\rev{One interesting trend is the relatively larger speedup of SC in the 8.3B model than in the 2.5B model.
Because the same number of GPUs are used, so the number of parameters per GPU, which affects the data-parallel gradient communication volume, becomes larger in the 8.3B model.
It increases the portion of the communication over the inter-stage communication.
Therefore, the speedup from compressing the data parallel communication in 8.3B becomes relatively larger than the 2.5B case.}

\cref{fig:pretrain} shows the curves of validation perplexity over the training of \rev{the 8.3B model}. 
With the use of \ppmethod (CB) and \embmethod (FE), the perplexity remains mostly the same compared to the baseline, and sometimes even performs better depending on which iteration the perplexity is measured. 
This is because \ppmethod successfully restricts the impact of compression errors to be resolved within the same iteration, especially using \lazy.
\Embmethod does not induce any mathematical changes to the baseline, and thus it is guaranteed to have no perplexity increase.
\Dpmethod provides a large amount of speedup at the cost of some perplexity trade-off.
Even with the error feedback techniques~\cite{powersgd}, it is inevitable that the error is applied after the update, which causes the staleness effect.
Nonetheless, \dpmethod controls the perplexity to have a marginal increase, while providing the most speedup.

We also conducted a language model validation of five zero-shot downstream tasks to validate our work based on~\cite{eval-harness}, as shown in~\cref{tab:zeroshot}.
\rev{Zero-shot tasks directly evaluate the pretrained model on some tasks (e.g., QnA) without fine-tuning and we used them to represent the expressibility of a model.}
For both the 8.3B and 2.5B models, CB and CB+FE show comparable accuracy on the baseline (no compression) model.
CB+FE+SC shows marginal accuracy degradation compared to baseline, which aligns with the trend of~\cref{tab:pretrain} and~\cref{fig:pretrain}.

\begin{figure}
    \centering
    \includegraphics[width=\columnwidth]{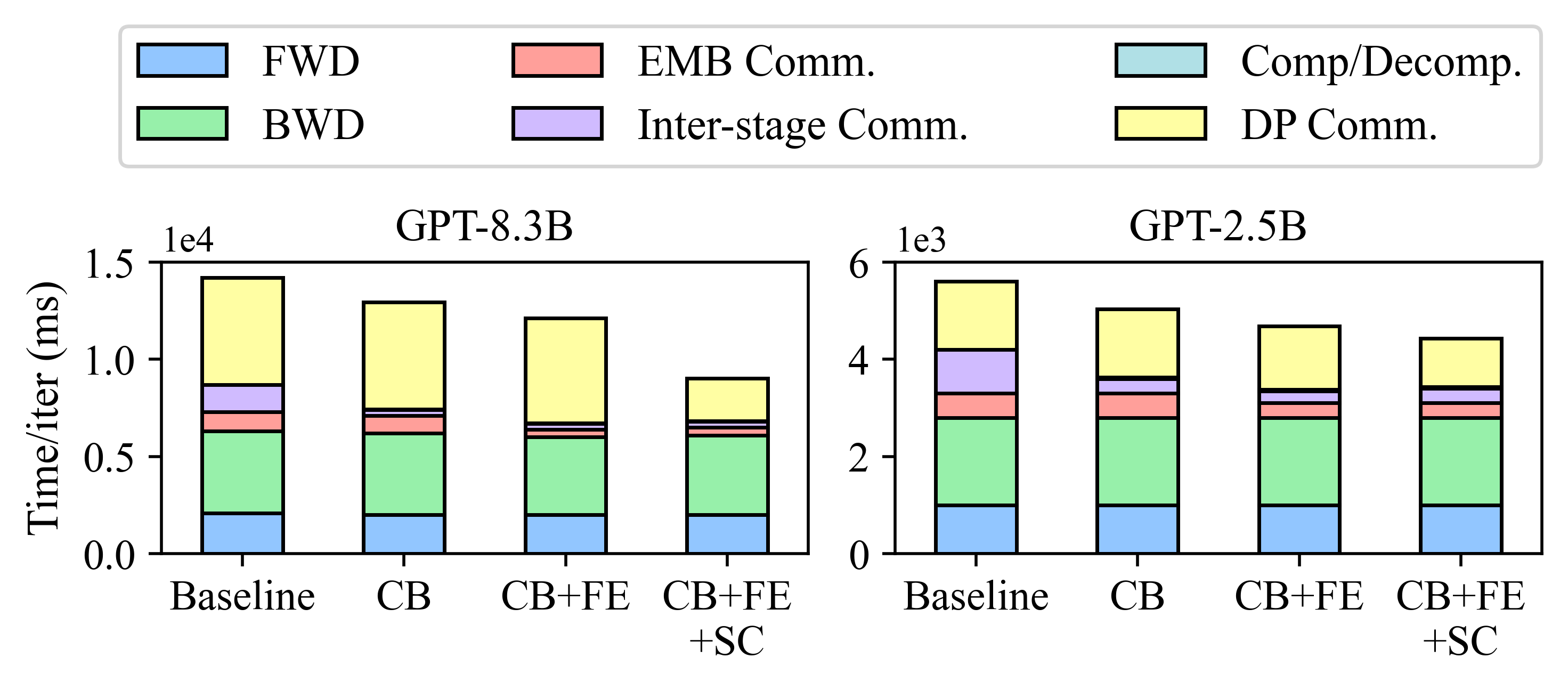}
    \caption{Breakdown of the execution times using 128 GPUs, in ablation of the proposed techniques.} 
    \label{fig:break}
\end{figure}

\cref{fig:break} shows the breakdown measured in the same way as in \cref{sec:moti}.
As depicted in the purple bars, \ppmethod (CB) reduces most of the backward inter-stage communications, by 78.57\% compared to the baseline.
Much of the inter-stage communication is left uncompressed because of \epi, but they are overlapped by the other computational stages and do not affect the training time.
Some portion of inter-stage communication that still remains in the stack accounts for the forward traffic which is not the target of \ppmethod.
The red bars represent the embedding synchronization part.
The reduction is about 40\%, which is almost identical to the analytic cost model provided in \cref{sec:embmethod} of 42.9\%.
When all the proposed methods have been applied (CB+FE+SC), the total communication time overhead has been reduced by 63.29\% in the 8.3B model, showing the effectiveness of \thiswork. 
Note that the compression and decompression overhead is negligible due to the extremely high throughput of the compression algorithm, which we will demonstrate in~\cref{sec:compdecomp}.

\subsection{Analysis of \PPmethod}
\label{sec:evallazy}

\begin{table}[t]
\centering
\caption{Effect of lazy error propagation on accuracies of zero-shot tasks in GPT-2.5B.}
\begin{tabular}{lccc}
\toprule
 \textbf{Tasks} & \textbf{Baseline} & \textbf{CB (Non-LEP)} & \textbf{CB (LEP)} \\ \midrule
 LAMBADA & \textbf{62.00\%} & \underline{61.79\%} & 61.93\% \\ 
 PIQA & \underline{71.76\%} &  71.87\% & \textbf{72.63\%} \\ 
 MathQA & 24.15\% & \underline{23.69\%}  & \textbf{24.25\%} \\
 WinoGrande & \textbf{62.19\%} & \underline{59.75\%} & 60.62\% \\
 RACE & 33.88\% & \underline{33.59\%} & \textbf{35.12\%} \\\bottomrule
\end{tabular}
\vspace{-3mm}
\label{tab:lep}
\end{table}

\cref{tab:lep} shows the effect of \lazy on the model quality.
`CB (Non-LEP)' refers to \ppmethod without \lazy, and `CB (LEP)' refers to \ppmethod with \lazy. 
\rev{\Epi was applied to all the cases because CB without \epi diverged.}
Bold accuracy is the highest, and underlined accuracy is the lowest.
While applying compression to the backpropagation without \lazy severely damages the model quality, which brings out the lowest accuracies, applying \lazy makes the model quality comparable to the baseline non-compressed model. 

\begin{figure}
    \centering
    \includegraphics[width=0.86\columnwidth]{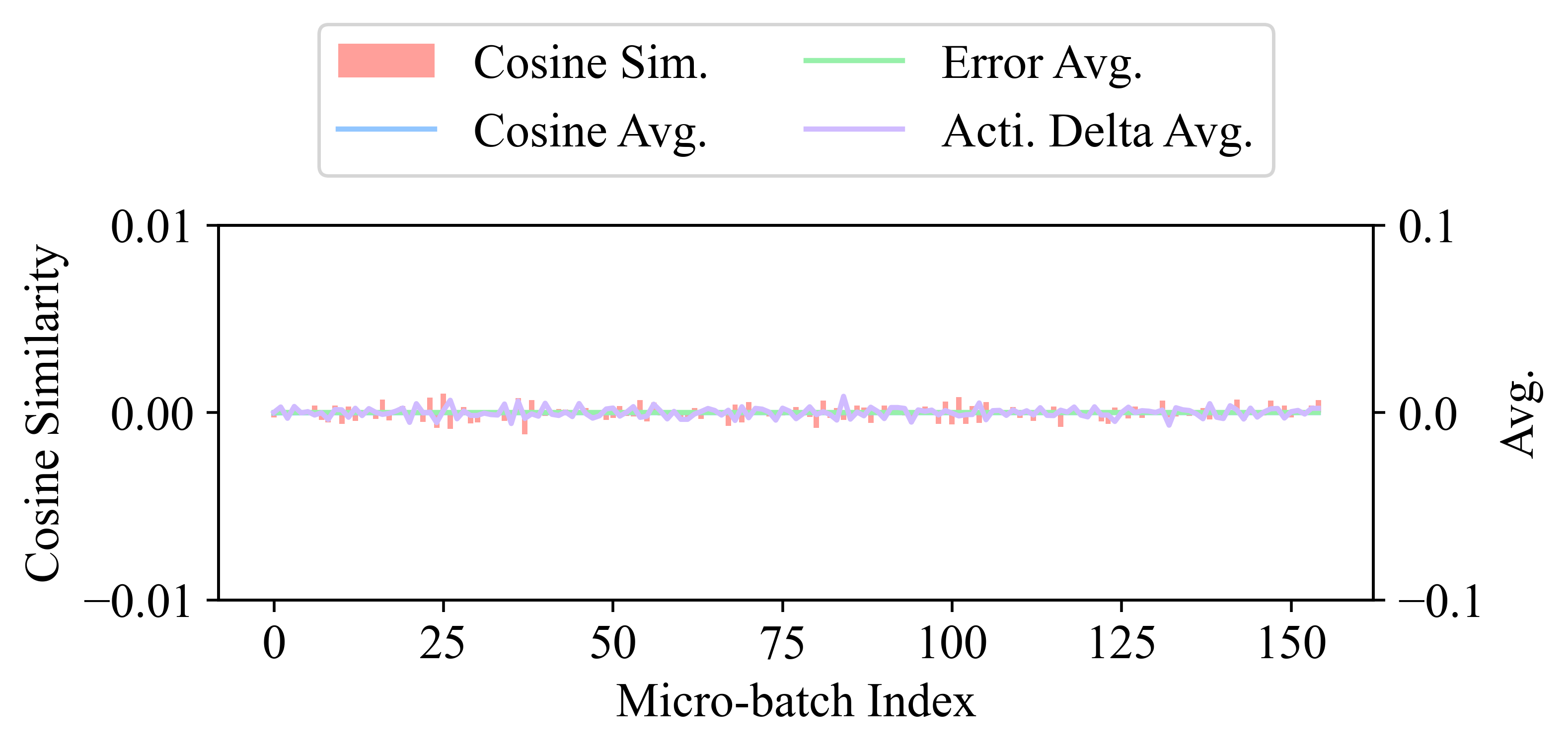}
    \caption{Cosine similarity of errors and activation differences.}
    \vspace{-3mm}
    \label{fig:lazy}
\end{figure}

\cref{fig:lazy} depicts how the conditions from \cref{eq:actcond} hold during training.
The green curves represent the average values for $\epsilon^{(i)}$ over 150 micro-batches during training.
In addition, the purple curves show that the average of the difference between activations ($Y^{(i)} - Y^{(i+n)}$) is also near zero.
Finally, the cosine similarity between $\epsilon^{(i)}$ and $Y^{(i)}-Y^{(i+n)}$ mostly stays around zero, which indicates that the two terms are independent.
This suffices that the conditions from \cref{eq:actcond} are true, leading $G^*$ in \cref{eq:gstar} to correctly approximate \cref{eq:g}.

\begin{figure}
    \centering
    \includegraphics[width=\columnwidth]{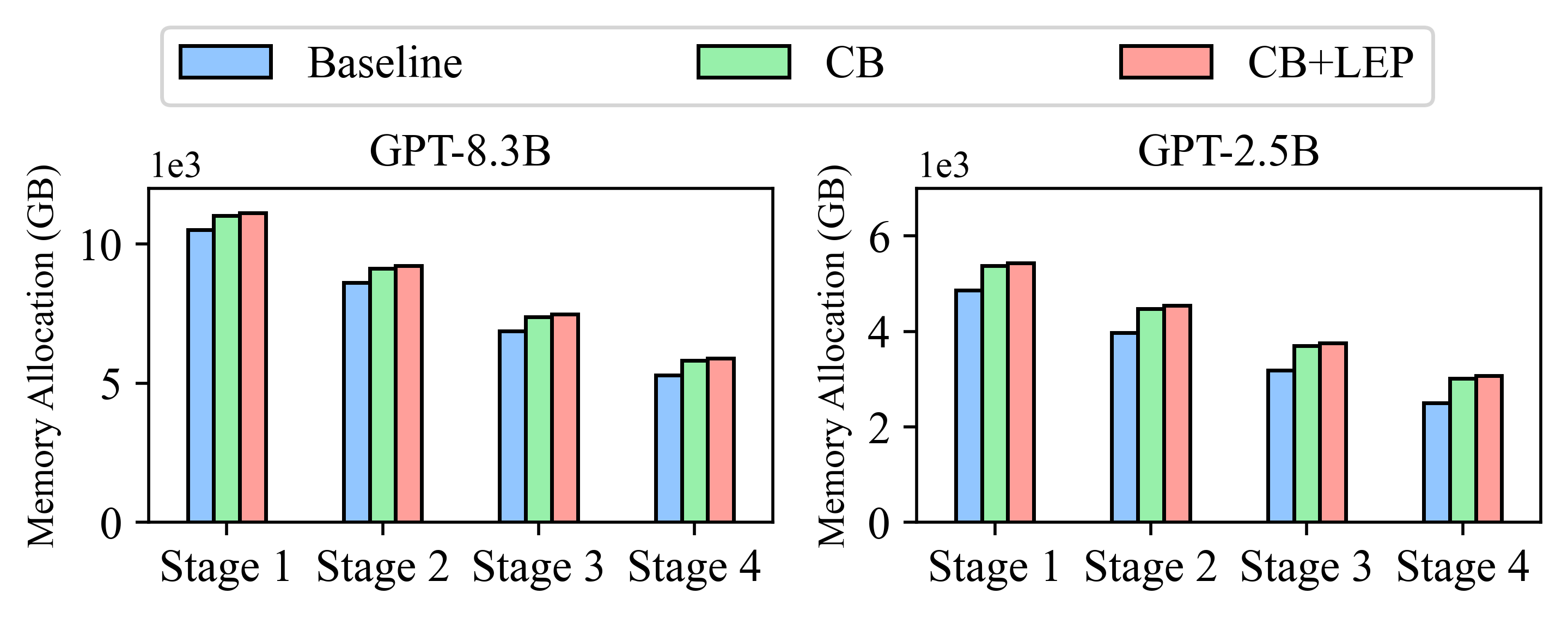}
    \caption{Maximum memory allocation per GPU of base compressed backpropagation and with lazy error propagation.}
    \label{fig:memory}
\end{figure}

\cref{fig:memory} shows the memory overhead of \ppmethod by plotting peak memory consumption reported by PyTorch.
To apply compression~\cite{powersgd}, a separate memory region has to be allocated for low-rank matrices. 
This accounts for the 5-10\% overhead to the baseline.
In addition, \lazy requires small additional memory for storing the error between micro-batches, but the overhead is marginal, which adds only 1\% additional overhead.

\subsection{Analysis of \DPmethod}
\label{sec:evalselcomp}

\begin{figure}
    \centering
    \includegraphics[width=\columnwidth]{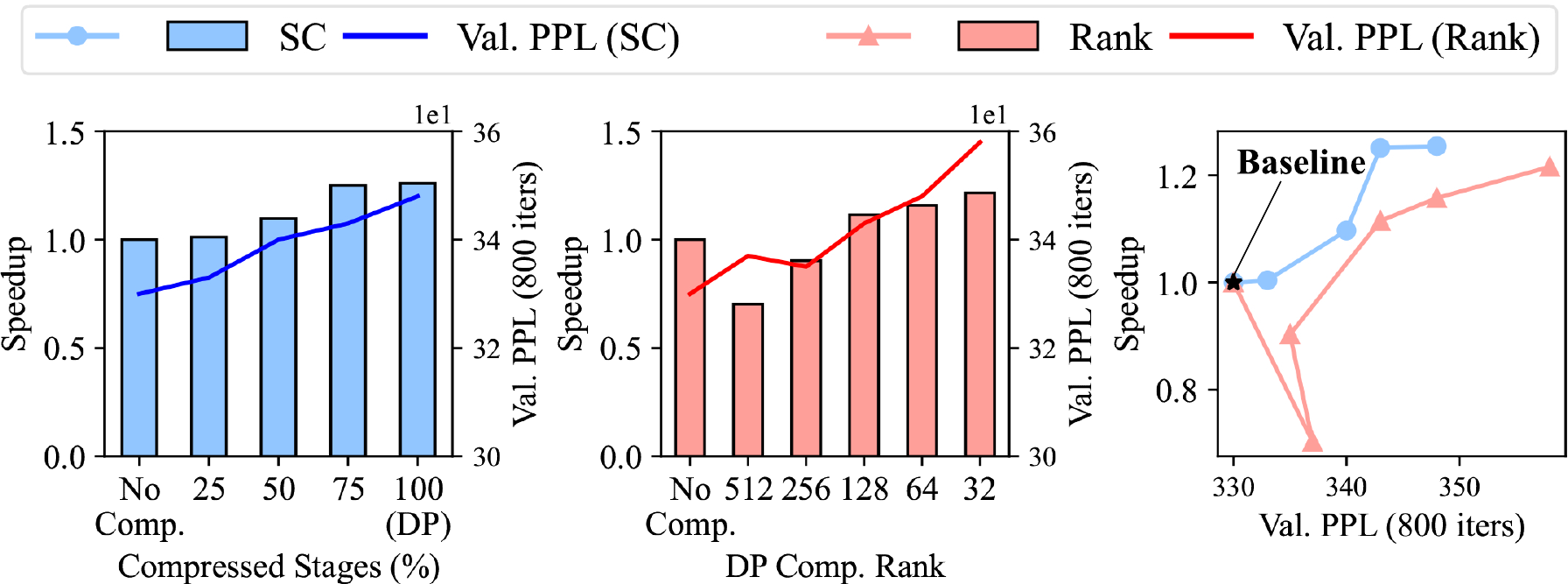}
    \caption{\rev{Effect of applying \dpmethod (left) and adjusting ranks (middle) to data-parallel communication on the speedup and the validation perplexity in GPT-2.5B.}} 
    \label{fig:selective}
\end{figure}

\rev{
One might wonder if the compression ratio (i.e., ranks) of the low-rank approximation can be adjusted instead of applying \dpmethod.
One important aspect is that the critical path cannot be considered by merely adjusting the compression ratio.
In this section, we show that \dpmethod provides a much better trade-off between the training speed and the model quality (measured with validation perplexity) on GPT-2.5B.

\cref{fig:selective} plots the training time and the perplexity of the two methods.
In the left figure, we apply \dpmethod and vary the percentage of stages being compressed.
With \dpmethod, we achieve a reasonable trade-off between the speedup and validation perplexity.
On the other hand, in the middle figure, we plot how the speedup and validation change by merely adjusting the rank used in the compression (compression ratio). Surprisingly, the relation between rank and perplexity is non-linear, which makes the traditional rank-adjusting infeasible as a tuning method.
At rank 512 which translates to about 10$\times$ compression rate, both the perplexity and speedup significantly worsen.
The reason for the speed degradation comes from the compression algorithm, where a too-high-valued rank will increase the time overhead for compression and decompression, as illustrated in~\cref{sec:compdecomp}.
Thus, relying on the compression ratio for the speed-accuracy trade-off would not be a rational choice.

\cref{fig:selective} (right) shows a direct comparison between \dpmethod and adjusting ranks by plotting the validation perplexity and the speedup together.
Considering that the upper-left (higher speedup and lower PPL) is the optimal direction, \dpmethod always provides a better trade-off than adjusting ranks.
We believe an even better trade-off can be achieved by automatically choosing the right combination of the compression rank and the number of stages for \dpmethod, which we leave as future work.
}




\subsection{Analysis on Configuration Sensitivity}
\label{sec:configsensi}

\begin{figure}
    \centering
    \includegraphics[width=0.9\columnwidth]{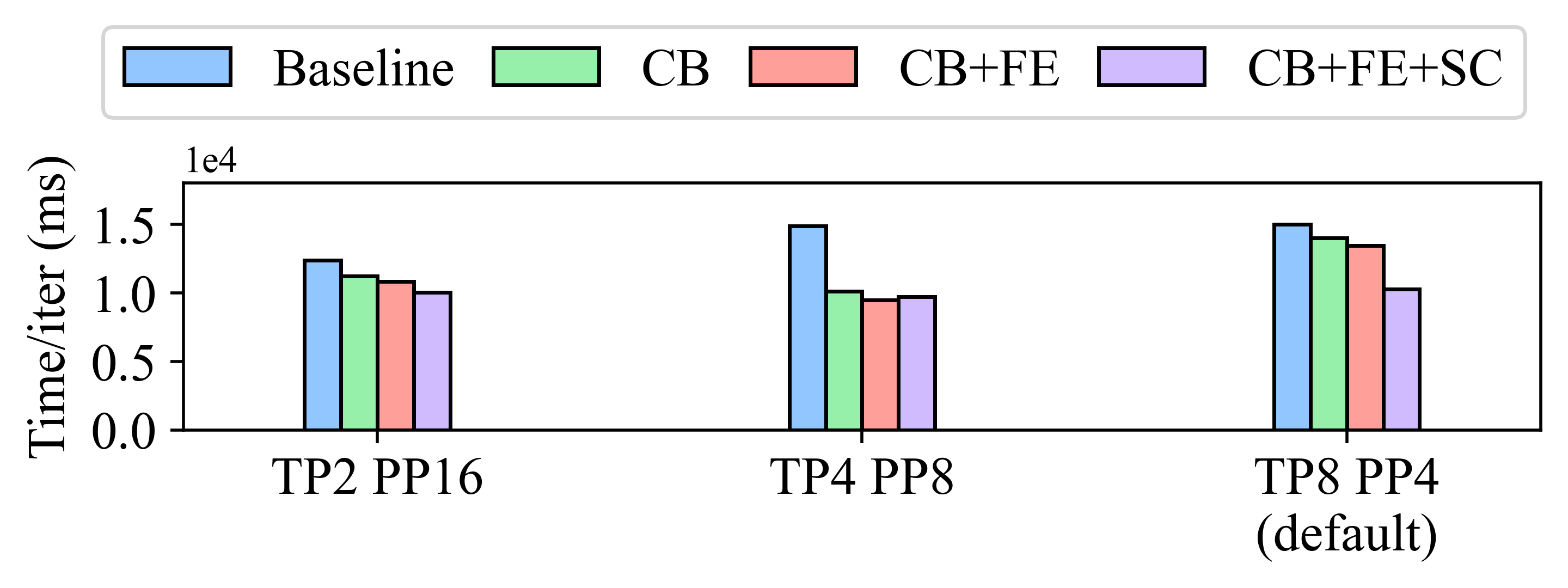}
     \caption{Tensor/pipeline-parallel configuration sensitivity of training time with the fixed data-parallel setting on GPT-9.2B.} 
    \label{fig:configsensi}
\end{figure}

Finding the best tensor/pipeline parallel configuration for training is an active research area~\cite{tarnawski2021piper}.
\cref{fig:configsensi} shows the training time of models on various parallel configurations.
We fixed the number of data-parallel ways to four for a fair comparison and conducted experiments on possible configuration settings.
We tested the configurations up to eight tensor-parallel ways because tensor-parallel ways are generally limited to the number of GPUs in a node.
With 128 GPUs, this results in the number of pipeline ways from 16 to 4.
To evenly divide the layers up to 16 stages for a fair comparison, we increased the number of layers to 80, which corresponds to 9.2B parameters.
For \dpmethod, we used the same 75\% compression for all settings.
\thiswork provides at least 19.2\% speedup for all parallel configurations.
The trend is that CB has more advantage when the number of pipeline-parallel ways increases because this incurs more inter-stage communication from deeper stages in the pipeline.
On the other hand, SC takes advantage as the number of pipeline-parallel ways decreases.
This is because, with less number of stages, the number of parameters per GPU increases, and thus the portion of data-parallel communication becomes larger.

\begin{figure}
    \centering
    \includegraphics[width=\columnwidth]{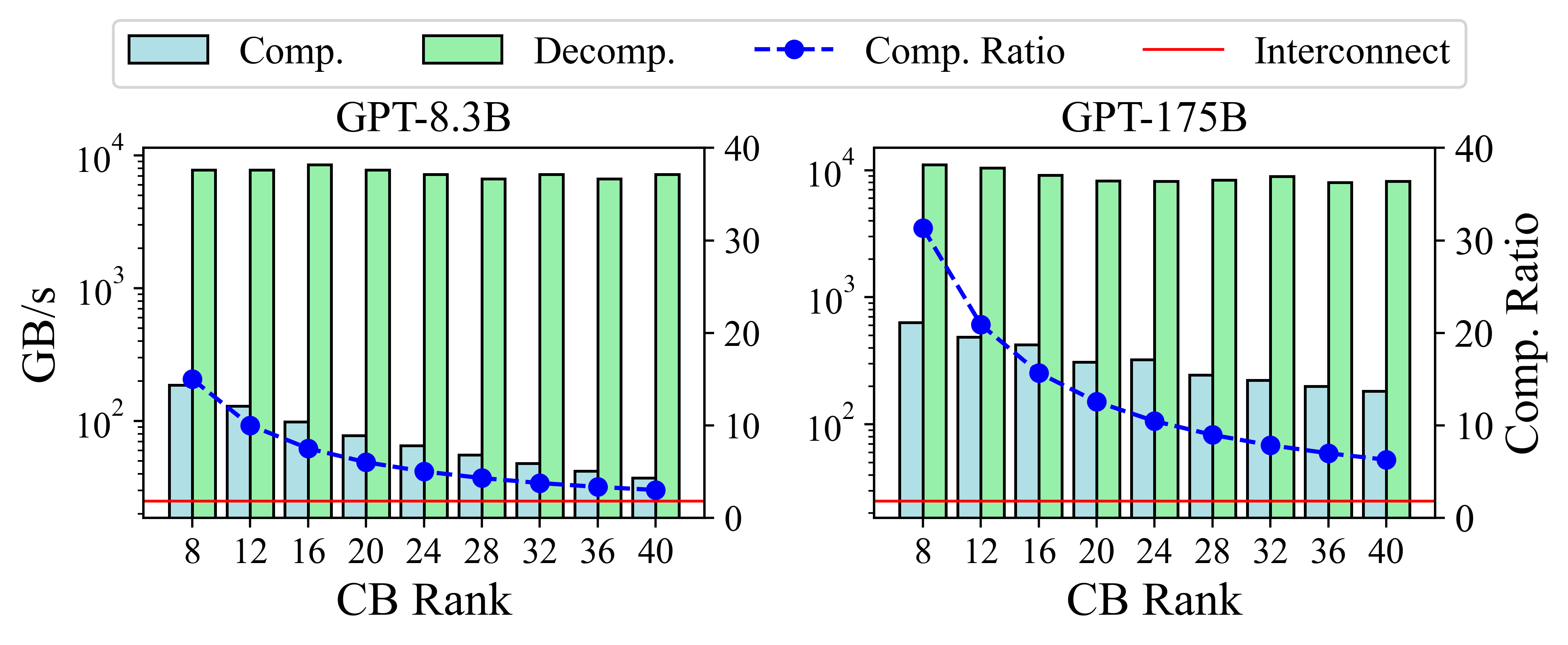}
    \caption{Throughput of inter-stage compression and decompression on GPT-8.3B (left) and GPT-175B (right).}
    \label{fig:throughput}
\end{figure}

\subsection{Analysis on Compression/Decompression Throughput}
\label{sec:compdecomp}

\thiswork uses a compression algorithm as its key component, and thus analyzing the compression and decompression throughput is critical.
In~\cref{fig:throughput}, we show that the compression and decompression throughput are much higher than that of the interconnect bandwidth.
In CB rank 16 of the 8.3B model, the compression throughput is 786.96Gbps (98.37GB/s), and the decompression throughput is 68.2Tbps (8.32TB/s), which has enough gap with the interconnection throughput of 200Gbps (25GB/s) depicted in red lines.

The compression throughput becomes higher in larger model sizes because constant setup overheads for the compression kernels become amortized with larger data.
An interesting and rather counter-intuitive trend is that the throughput decreases with higher CB ranks (less compression).
This is because the orthogonalization phase is the main bottleneck (about 80\%) for the compression algorithm, which takes longer with a larger output size. 

\subsection{Analysis on Scalability}
\label{sec:scalability}

\begin{figure}
    \centering
    \includegraphics[width=0.9\columnwidth]{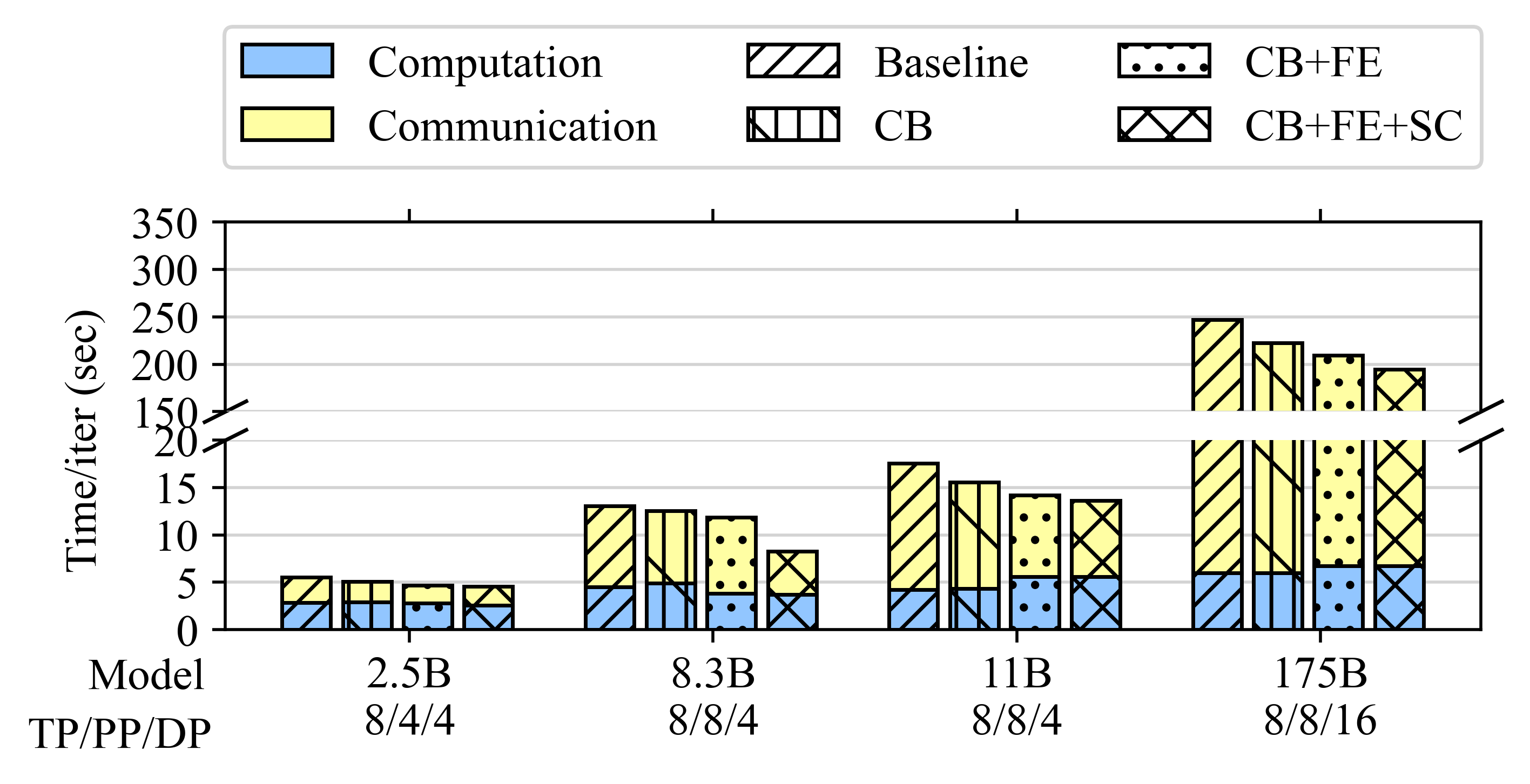}
    \caption{Scalability of our mechanisms.} 
    \label{fig:scalability}
\end{figure}

\cref{fig:scalability} shows the scalability of the proposed work with four Megatron-LM~\cite{megatron-lm} based models.
We fixed tensor-parallel-ways to 8 and increased the number of GPUs in larger models for a fair comparison.
\thiswork scales well on larger model sizes even when the model grows up to 175B (GPT-3)~\cite{brown2020language}.

The scalability comes from two factors. 
First, it is well-known that larger models suffer more from communication overheads~\cite{smith2022using}; there is more potential for the proposed work.
Second, as shown in \cref{sec:compdecomp}, the compression itself becomes more efficient when the model size becomes larger. 
The compression overhead was already small for 2.5B and 8.3B models, but becomes even smaller with extremely larger models.

\section{Discussion}

\subsection{Application on Other Accelerators}
Aside from GPUs, we can use other DL accelerators for training large-scale models, such as TPU~\cite{tpu} and IPU~\cite{graphcoreipu}.
They generally have higher computational throughputs and intra-node bandwidth than GPUs, and their inter-node speed (400Gbps for TPU and 100Gbps for IPU) is similar to that of GPUs.
They also use 3D parallelism for large-scale training, which requires inter-node communication.

\thiswork will have more potential on these accelerators because computational throughput over inter-node bandwidth is larger than our setting.
For an example of IPU-POD128, it provides 8 PFLOPS per node, while our setting provides 5 PFLOPS per node.
However, its inter-node communication is 100Gbps, which is half the bandwidth of our setting.
In such a case, \thiswork will provide more advantages.

\subsection{Application on Other Models}
\rev{
While \textit{3D Parallelism} is widely used to train NLP models, it can also be applied to other domains (e.g., CNN).
\thiswork can be adopted to these models, because the mechanisms of the proposed techniques are independent to a model structure.

For example, training of AmoebaNet~\cite{amoebanet} is often done with 3D Parallelism~\cite{gpipe} to mitigate the problem of increased model size and training time.
However, the parallelism makes the inter-stage and data-parallel gradient communication overhead significant.
Gradient compression methods can reduce the overhead of the data-parallel gradient communication, but suffer from an accuracy degradation problem.
In this circumstance, \thiswork can minimize the accuracy degradation by \textit{\dpmethod} and further accelerate the training by compressing the inter-stage communication through \textit{\ppmethod}.

In fact, we believe \thiswork can be applied to any DNN domain that requires a model larger than a single device. 
For example, modern graph neural networks started adopting data parallelism~\cite{dgl,roc} and pipeline parallelism~\cite{pipegcn}.
\thiswork could be applied to such cases to bring a speedup.
}

\section{Related Work}

\subsection{Data Parallelism}
\label{sec:Dataparallel}
To cope with the growing size of the large-scale models~\cite{efficientNet, bert}, plenty of methods were proposed to accelerate the training procedure. 
\emph{Data parallelism}~\cite{trainingImagenetInHour,largeScaleDistributed, scaleOutLargeMinibatchSGD, elasticAveragingSGD} has been commonly used to accelerate model training in a distributed manner.
Data parallelism is to copy the entire model to every worker and train with a different mini-batch while keeping an identical weight among all the workers. 
In order to keep exactly the same state after every step, data parallelism necessitates the synchronization of gradients.
Its communication overhead grows linearly proportional to the model size which makes all-reduce~\cite{optimizationOfCollectiveComm, trainingImagenetInHour} be the main bottleneck of training.
To mitigate this problem, many researchers tried to optimize communication itself~\cite{poseidon, flexreduce, blueconnect, blink,canary, adasum} while maximizing the overlap between communication and computation~\cite{priorityBasedParameterProp, byteScheduler, tictac, breaking}.

Another effective approach to reducing communication overhead in data parallelism is \emph{gradient compression}. By using a low-rank approximation of gradient matrices~\cite{powersgd, wang2021pufferfish,cho2019gradzip}, sparse gradient update methods~\cite{bernstein2018signsgd, dgc, scalecom, MLSYS2021_6ea9ab1b, MLSYS2021_8613985e} and gradient quantization~\cite{faghri2020adaptive, basu2019qsparse, yu2019double}, communication volume can be effectively reduced without significant loss in accuracy.
To prevent further performance degradation caused by gradient compression, most works adopt \emph{error-feedback}
which is to compensate for the difference between the compressed gradient and the original one. 
Recently, statistical ways~\cite{qian2021error, horvath2020better, xu2021step} can be alternatively used to avoid additional computation overhead when conducting compression methods.

\subsection{Tensor and Pipeline Parallelism}
\label{sec:Modelparallel}
The size of transformer-based language models~\cite{vaswani2017attention, t5, bert,brown2020language} has been grown
at an exponential rate~\cite{sc21efficient}. 
This trend hits on limited accelerator memory capacity in addition to proportional growth of training time~\cite{brown2020language}. 
However, the aforementioned data parallelism cannot handle both issues by scaling larger batch size~\cite{lin2018don}. Tensor and pipeline parallelism has tried to handle both issues in their own ways.


\emph{Tensor parallelism}~\cite{ wang2018unifying, jia2019beyond,dryden2019channel,tarnawski2021piper} is to hand out parameters (tensor) in the same layer to multiple workers. 
Tensor parallelism is mainly focused on reducing the number of synchronization points among workers sharing the same layer. 
On the other hand, \emph{pipeline parallelism}~\cite{gpipe} schedules the execution of \emph{micro-batches} which are sampled from a mini-batch. 
A primary constraint of pipelined schedule is \emph{synchronous execution} which is essential regulation in the model training process. The activation calculated in the forward pass has to be used in the corresponding backward pass. In such regard, recent works on pipeline parallelism have mainly focused on reducing memory overhead~\cite{rajbhandari2020zero,  rajbhandari2021zero, ren2021zero} for handling \emph{staleness problem}~\cite{ pipemare,pipescale} while reducing \emph{pipeline bubble}~\cite{ pipedream, fan2021dapple, li2021chimera} by optimizing pipelined schedule. 

Many transformer-based models can be successfully trained by using a combination of three distributed training methods which is called \emph{3D parallelism}~\cite{megatron-lm, dpPlusMp, hypar, sc21efficient}. However, to the extent of our knowledge, no attempt has been made to apply 3D parallelism-aware communication compression to a large language model. Before \thiswork, the effect of compressed communication in 3D parallelism has been unknown space.


\section{Conclusion}
In this work, we proposed \thiswork, which compresses the communications of large, distributed NLP models that utilize 3D parallelism. 
Because the conventional communication compression algorithms fail to exploit pipeline-related opportunities and result in a model quality drop, we proposed multiple techniques that reduce the amount of communications while maintaining the model quality.
We believe the impact of \thiswork will be more significant with even larger models, adding value to the work in the upcoming future.

\begin{acks}
This work was supported by 
Samsung Electronics Co., Ltd (IO210809-08878-01), 
the National Research Foundation of Korea (NRF) grants (2022R1C1C1011307, 2022R1C1C1008131) and
Institute of Information \& Communications Technology Planning \& Evaluation (IITP) grant funded by the Korea government (MSIT) (2021-0-00853, 2020-0-01361).
Jaeyong Song and Youngsok Kim are with the Department of Computer Science at Yonsei University and are partly supported by the BK21 FOUR (Fostering Outstanding Universities for Research) funded by the Ministry of Education (MOE) of Korea and National Research Foundation (NRF) of Korea.
\end{acks}

\section*{Data Availability Statement}
The artifact of~\thiswork is available at~\cite{optcc}.
It contains NLP dataset generation code, \cite{powersgd}-based compression code, \cite{megatrongithub}-based training code and \cite{eval-harness}-based evaluation code. 

\balance
\bibliographystyle{ACM-Reference-Format}
\bibliography{references}










\end{document}

%% file: packages.tex
\usepackage[htt]{hyphenat}
\usepackage{amsmath,amsfonts}
\usepackage{algorithmic}
\usepackage{textcomp}
\usepackage{xcolor}
\usepackage{nicefrac}
\usepackage{tabularx}
\newcolumntype{Y}{>{\centering\arraybackslash}X}

\usepackage{subcaption}

\usepackage{enumitem} 
\usepackage{tcolorbox}

\usepackage[capitalize]{cleveref}

\definecolor{olivegreen}{rgb}{0, 0.6, 0}
\definecolor{redorange}{HTML}{FF5349}
\definecolor{blue(ncs)}{rgb}{0.0, 0.53, 0.74}
\definecolor{navy}{HTML}{273BE2}

\definecolor{black}{HTML}{000000}
\definecolor{white}{HTML}{ffffff}
\definecolor{color1}{HTML}{ACE5EE}
\definecolor{color2}{HTML}{0093AF}
\definecolor{color3}{HTML}{CC0000}
\definecolor{color4}{HTML}{0087BD}
\definecolor{color5}{HTML}{333399}
\definecolor{color6}{HTML}{20B2AA}

\usepackage{xspace}         
\usepackage{graphicx}

\usepackage{pgfplots}
\usepackage{pgfplotstable}
\usepgfplotslibrary{groupplots}

\usepackage{array, makecell} %
\newcolumntype{x}[1]{>{\centering\arraybackslash\hspace{0pt}}p{#1}}

\usepackage{multirow}

\usepackage{pifont}

\usepackage{booktabs}
\usepackage{balance}

%% file: commands.tex
\newcommand{\rev}[1]{#1}

\newcommand{\JL}[1]{{\color{olivegreen}[\textbf{\sc JLee}: \textit{#1}]}}
\newcommand{\JW}[1]{{\color{redorange}[\textbf{\sc JJung}: \textit{#1}]}}
\newcommand{\JY}[1]{{\color{blue(ncs)}[\textbf{\sc JSong}: \textit{#1}]}}
\newcommand{\JK}[1]{{\color{magenta}[\textbf{\sc JKYim}: \textit{#1}]}}
\newcommand{\HS}[1]{{\color{navy}[\textbf{\sc HJang}: \textit{#1}]}}

\renewcommand{\JL}[1]{}
\renewcommand{\JW}[1]{}
\renewcommand{\JY}[1]{}
\renewcommand{\JK}[1]{}
\renewcommand{\HS}[1]{}

\newcommand{\thiswork}{Optimus-CC\xspace} 

\newcommand{\ppmethod}{compressed backpropagation\xspace}
\newcommand{\PPmethod}{Compressed Backpropagation\xspace} 

\newcommand{\epi}{epilogue-only compression\xspace}
\newcommand{\Epi}{Epilogue-only compression\xspace}

\newcommand{\lazy}{lazy error propagation\xspace}
\newcommand{\Lazy}{Lazy error propagation\xspace} 
\newcommand{\LAZY}{Lazy Error Propagation\xspace} 

\newcommand{\embmethod}{fused embedding synchronization\xspace}
\newcommand{\Embmethod}{Fused embedding synchronization\xspace} 
\newcommand{\EMBmethod}{Fused Embedding Synchronization\xspace} 

\newcommand{\dpmethod}{selective stage compression\xspace}
\newcommand{\Dpmethod}{Selective stage compression\xspace} 
\newcommand{\DPmethod}{Selective Stage Compression\xspace} 

\newcommand\independent{\protect\mathpalette{\protect\independenT}{\perp}}
\def\independenT#1#2{\mathrel{\rlap{$#1#2$}\mkern2mu{#1#2}}}

\newcommand*\circled[1]{\tikz[baseline=(char.base)]{
            \node[shape=circle,draw,inner sep=0.4pt] (char) {#1};}}